\documentclass{article} 

\usepackage[usenames,dvipsnames]{xcolor}

\usepackage{geometry}

\usepackage{times}

\usepackage{amsmath,amsfonts,bm}

\def\eqref#1{equation~\ref{#1}}

\def\1{\bm{1}}

\DeclareMathAlphabet{\mathsfit}{\encodingdefault}{\sfdefault}{m}{sl}
\SetMathAlphabet{\mathsfit}{bold}{\encodingdefault}{\sfdefault}{bx}{n}

\DeclareMathOperator*{\argmax}{arg\,max}

\usepackage{url}            
\usepackage{booktabs}       
\usepackage{amsfonts}       
\usepackage{nicefrac}       
\usepackage{float}
\newfloat{myframe}{htbp}{lof}[figure]
\floatname{myframe}{Figure}
\definecolor{mydarkblue}{RGB}{0,0,139}

\usepackage[most]{tcolorbox}
\newtcolorbox{exampleenv}{
width=1.0\linewidth,
left=0mm,
right=0mm,
top=0mm,
bottom=0mm,
halign=center,
colframe=black,
boxrule=0.5pt,
fontupper=\tiny
}
\newtcolorbox{exampleenvborder}{
center,
top=-1mm,
bottom=-1mm,
right=0mm,
left=0mm,
width=1.0\linewidth,
halign=center,
colframe=black,
colback=white,
boxrule=1.0pt,
}

\usepackage{tabularx}
\usepackage{makecell}
\usepackage{natbib}
\usepackage{listings}
\usepackage{float}
\usepackage{siunitx}
\usepackage{subcaption}
\usepackage{graphicx}
\usepackage[export]{adjustbox}
\usepackage[inline]{enumitem}
\usepackage{mdframed}
\mdfdefinestyle{splitstyle}{
backgroundcolor=gray!20
}
\usepackage{multirow}

\usepackage{amsmath,amssymb,amsthm}
\usepackage[mathscr]{eucal}
\usepackage{stmaryrd}
\usepackage{accents}
\usepackage{upgreek}
\usepackage[scaled=0.8]{FiraMono}

\usepackage{booktabs,colortbl,multirow}

\usepackage{pifont}

\usepackage{algorithm,algorithmic}
\usepackage{wrapfig}

\graphicspath{{figs//}}

\allowdisplaybreaks

\newcommand{\witta}[1]{\textcolor{Mulberry!80!black}{[\hl{\textbf{WJ}}: #1]}}

\newcommand{\todoakmInline}[1]{\todo[inline,color={RoyalBlue!35!white}]{#1}}

\newcommand{\removefornow}[1]{}

\renewcommand{\Pr}{\mathbb{P}}

\newcommand{\XCal}{\mathscr{X}}
\newcommand{\YCal}{\mathscr{Y}}

\newcommand{\defEq}{\stackrel{.}{=}}

\newcommand{\Real}{\mathbb{R}}

\usepackage[colorlinks=true,citecolor=ForestGreen]{hyperref}
\usepackage{url}
\usepackage[toc,page,header]{appendix}
\usepackage{soul}

\usepackage[textsize=tiny,textwidth=2cm]{todonotes}

\usepackage{tcolorbox}

\tcbset{width=\textwidth,boxrule=0pt,colback=gray!25,arc=0pt,auto outer arc,left=0pt,right=0pt,boxsep=5pt}

\newcommand{\ourtitle}{Language Model Cascades: \\ Token-level uncertainty and beyond}
\title{\ourtitle}

\author{Neha Gupta\footnote{Corresponding author: \texttt{nehagup@google.com}} \qquad Harikrishna Narasimhan \qquad Wittawat Jitkrittum \\
Ankit Singh Rawat \qquad Aditya Krishna Menon \qquad Sanjiv Kumar \\[2em]
Google Research, New York\\
}

\usepackage[toc,page,header]{appendix}
\usepackage{minitoc}

\begin{document}

\maketitle

\doparttoc 
\faketableofcontents 

\begin{abstract}
Recent advances in language models (LMs) have led to significant improvements in quality on complex NLP tasks,
but at the expense of increased inference costs.
\emph{Cascading} offers a simple strategy to achieve more favorable cost-quality tradeoffs:
here, 
a small model is invoked for most ``easy'' instances, while 
a few ``hard'' instances are deferred to the large model.
While the principles underpinning cascading are well-studied for classification tasks
---
with deferral based on predicted class uncertainty favored theoretically and practically
---
a similar understanding is lacking for generative LM tasks.
In this work, we initiate a systematic study of deferral rules for LM cascades.
We begin by examining
the natural extension of predicted class uncertainty to generative LM tasks,
namely,
the predicted \emph{sequence} uncertainty. 
We show that this measure suffers from the \emph{length bias} problem, either over- or under-emphasizing outputs based on their lengths. 
This is because LMs produce 
a \emph{sequence} of uncertainty values,
one for each output token; 
and moreover, the number of output tokens is variable across examples.
To mitigate this issue, we propose to exploit the richer token-level uncertainty information implicit in generative LMs. 
We argue that na\"{i}ve predicted sequence uncertainty corresponds to a simple aggregation of these uncertainties. 
By contrast, we show that incorporating {token-level} uncertainty through learned \emph{post-hoc deferral rules} can significantly outperform such simple aggregation strategies, via experiments on a range of natural language benchmarks with FLAN-T5 models.
We further show that incorporating embeddings from the smaller model and intermediate layers of the larger model can give an additional boost in the overall cost-quality tradeoff.

\end{abstract}

\section{Introduction}
Recent advances in generative language modeling have yielded a series of 
Transformer-based models 
with remarkably improved \emph{quality} on complex NLP tasks~\citep{Radford:2018,Raffel:2020,Brown:2020,Black:2022,Hoffmann:2022,Chowdhery:2022,Wei:2022,Chung:2022,Tay:2023,Anil:2023,Touvron:2023,Gemini:2023}.
Unfortunately, such models also involve significantly increased \emph{inference costs},
which has motivated a series of efforts at reducing the same.
These span 
careful infrastructure optimization~\citep{Chowdhery:2022,Pope:2022,Sheng:2023},
rethinking the autoregressive decoding that underpin Transformers~\citep{Stern:2018,Leviathan:2023,Chen:2023,Sun:2023},
modifications of the underlying model architecture~\citep{Dao:2022}, and
 model compression strategies~\citep{Frantar:2023,Agarwal:2023}.

\emph{Cascading} is one simple strategy to achieve more favorable cost-quality tradeoffs
via \emph{adaptive} inference.
In a two-model cascade,
a small model is invoked for most ``easy'' instances, while 
a few ``hard'' instances
are deferred to
a large model.
Cascades have been widely explored in the vision domain~\citep{Viola:2001,Trapeznikov:2013,Bolukbasi:2017,Huang:2018,Rawat:2021,Kag:2023,Jitkrittum:2023},
and have seen increasing adoption within NLP~\citep{Mamou:2022,Varshney:2022,Khalili:2022,Dohan:2022,Chen:2023a,Chen:2023}.
Importantly, cascades can be implemented in a black-box fashion over existing models, and do not necessitate any additional training. 

\begin{figure}[!t]
    \centering
    \centerline{
    \includegraphics[width=1.0\linewidth]{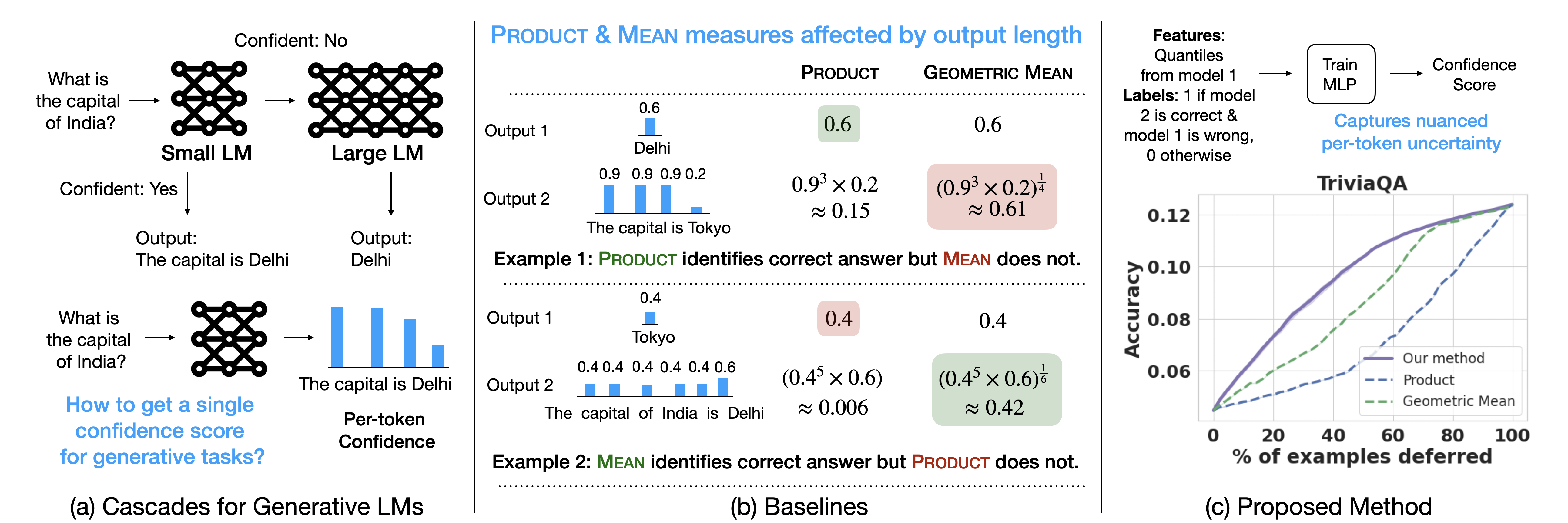}
    }
    \caption{{\bf (a)} In cascades, small models are used for easy instances whereas hard instances are routed to larger models. For generative LMs, the key challenge is to design a deferral rule based on uncertainties from multiple tokens. {\bf (b)} Standard baselines which take the product and geometric mean of the probabilities are affected by the length of the output and perform sub-optimally. {\bf (c)} Our proposed solution captures nuanced per-token uncertainty and outperforms both baselines.
     }
    \label{fig:main_figure}
\end{figure}
The key challenge in cascading is to design a \emph{deferral rule} which decides whether to defer an input to the larger model. The principles underpinning optimal deferral rules are well known for the classification setting, where the standard recipe is to employ the small model's prediction \emph{confidence},
as canonically measured by its softmax probability output (\emph{Chow's rule}~\citep{Cho1970}). This simple deferral rule is remarkably hard to surpass in most natural settings~\citep{Jitkrittum:2023}.

However, the narrative is more complex for generative LMs.
While one can na\"{i}vely translate Chow's rule for such models
based on the softmax probability of the output sequence,
this suffers from a
\emph{length bias} issue: 
one tends to defer longer predictions, regardless of quality.
Further, simply normalizing the probability by sequence length tends to over-correct this bias, and defer shorter predictions.
Intuitively, such na\"{i}ve translations of Chow's rule 
ignore a key distinction between the classification and generative LM setting:
the former involves a \emph{single} probability distribution over labels,
while
the latter involves a \emph{sequence} of distributions over multiple tokens of the LM output;
moreover, the number of output tokens differs across examples.
This variability 
complicates summarizing the sequence of uncertainty (or confidence) values into a single deferral score.


To mitigate the length bias and capture fine-grained information from the uncertainty vector over tokens, we propose to use \emph{quantiles} over the vector. 
Via experiments on a range of NLP benchmarks and FLAN-T5 models, we show that these quantiles can capture rich and complementary sources of uncertainty information from the uncertainty sequence vector and do better than the simple aggregation schemes like sum and average. 
However, we observe that there is no fixed quantile value which works across all datasets. This motivates us to \emph{learn} a deferral rule based on these quantile values as features, which can combine the strengths of these different quantile scores.
We show that our trained deferral rule is the most consistently performant method compared to all the natural baseline 
aggregation strategies. 
We further show that using embeddings from the smaller model and intermediate embeddings from the larger model can give further performance improvement.

To summarize, our contributions are:
\begin{enumerate}[label=(\roman*),itemsep=0pt,topsep=0pt,leftmargin=16pt]
    \item We show that simple sequence-level LM confidence measures for deferral 
    can yield strongly sub-optimal cost-quality tradeoffs,
    owing to a length bias issue (\S\ref{sec:chow-sum-bad}).

    \item We introduce \emph{token-level uncertainty} in the form of distribution quantiles, 
    and show that they 
    can yield to consistently more effective cost-quality tradeoffs,
    owing to their finer-grained measure of uncertainty.
    However, there is no \emph{fixed} quantile which works well across all settings (\S\ref{sec:chow-sum-bad}).

    \item We propose a post-hoc deferral rule trained on quantile features, and show it can outperform all other strategies on a range of NLP benchmarks for FLAN-T5 models
    (\S\ref{sec:router_experiments}).
    %
    We further demonstrate that 
    using 
    the large model's \emph{intermediate embeddings}
    can significantly
    boost
    performance. 
\end{enumerate}



%
\section{Background and Problem Setup}
\label{sec:background-cascades}

In this section, we discuss the relevant background and set up the problem of 
LM cascades. 

\textbf{Language models (LMs)}.~Given a finite vocabulary $\mathscr{V}$
(e.g., tokens derived from SentencePiece~\citep{Kudo:2018}),
a \emph{language model} (\emph{LM})
defines a distribution $p( \cdot \mid \boldsymbol{x} ) \in \Delta( \mathscr{V} )$ over all possible tokens given any \emph{context} $\boldsymbol{x} = ( x_1, \ldots, x_m ) \in \mathscr{V}^m$.
This in turn defines a distribution over \emph{sequences} $\boldsymbol{y} = ( y_1, \ldots, y_n ) \in \mathscr{V}^n$ for any $n \in \mathbb{N}_+$, with 
$p( y_1, \ldots, y_n \mid \boldsymbol{x} ) = p( y_1 \mid \boldsymbol{x} ) \cdot \prod_{i = 1}^{n - 1} p( y_{i + 1} \mid \boldsymbol{x}, y_1, \ldots, y_{i} )$ via the chain rule of probability.

LMs based on Transformers~\citep{Vaswani:2017} have proven increasingly popular in recent years.
Such models are typically \emph{pre-trained} on large corpora based on self-supervised objectives~\citep{Radford:2018,Devlin:2019,Raffel:2020,Brown:2020,Tay:2022,Anil:2023}.
These objectives
involve different (input, output) pair constructions
$( \boldsymbol{x}, \boldsymbol{y} )$
(e.g., masking out the next token),
upon which one minimizes the \emph{cross-entropy} or log-loss, i.e., ${-\log p( \boldsymbol{y} \mid \boldsymbol{x} )}$.

At inference time, given a trained LM and any input context $\boldsymbol{x}$, 
it is common to perform either \emph{classification} or \emph{generation}.
In the former, given a set of predefined choices $\mathscr{C} = \{ \boldsymbol{c}_i \}_{i \in [L]}$ (e.g., \{ {\tt yes}, {\tt no} \}), 
one scores each $p( \boldsymbol{c}_i \mid \boldsymbol{x} )$ and returns the highest scoring choice.
In the latter, 
one performs \emph{sampling} from $p( \cdot \mid \boldsymbol{x} )$ to produce a suitable output string response, e.g., by 
temperature~\citep{Ficler:2017},
top-$k$~\citep{Fan:2018}, 
or nucleus sampling~\citep{Holtzman:2020}.

%

%

\textbf{Model cascades}.~Cascades are a simple, generic strategy to improve the inference cost-quality tradeoff~\citep{wang2022wisdom}.
Given a collection of models of varying inference cost,
the key idea is to perform
\emph{adaptive} inference:
``easy'' samples are afforded less computation compared to ``hard'' samples.
Concretely, for any test input, 
one first executes the lowest cost model, 
and uses a \emph{deferral rule} to determine whether 
to terminate with its prediction, or 
to invoke the next cheapest model.
Cascades can reduce the average inference cost if only a small fraction of inputs are deferred.

Cascades
have a long history of usage in vision~\citep{Viola:2001,Huang:2018,Wang:2018},
where they are 
often applied for \emph{classification} problems.
Given an instance space $\mathscr{X}$ and label space $\mathscr{Y}$, the classification problem seeks a \emph{classifier} 
$h \colon \XCal \to \YCal$
with good \emph{average quality} under some distribution $\Pr$,
as measured by 
$\mathbb{E}_{(x, y) \sim \Pr}[ q( x, y, h( x ) ) ]$
for 
some $q( x, y, h( x ) ) \in \Real_+$.
In the simplest case, $q( x, y, \hat{y} ) = \1( y = \hat{y} )$ measures the classifier \emph{accuracy}.

Now suppose we have two classifiers $h^{(1)}, h^{(2)}$, 
with inference costs (e.g., latencies) 
$c^{(1)} \ll c^{(2)}$.
Operationally, a cascade first invokes the ``small'' model $h^{(1)}$, and then applies a deferral rule to decide whether to either \emph{defer} prediction to the ``large'' model $h^{(2)}$, or terminate with $h^{(1)}$'s prediction.
More precisely,
let
$r \colon \mathscr{\XCal} \to \{ 0, 1 \}$ denote the deferral rule,
where 
$r( x ) = 1$ denotes that we wish to defer to the large model.
Then, the cascaded classifier is~\citep{Kag:2023,Jitkrittum:2023}:
$$ h^{\rm cas}( {x} ) = 
\1( r( {x} ) = 0 ) \cdot h^{(1)}( {x} ) + \1( r( {x} ) = 1 ) \cdot h^{(2)}( {x} ). $$
Given an input ${x}$,
the corresponding cascade quality and cost are:
\begin{align*}
    Q( {x}, y, h^{\rm cas}( {x} ) ) &\defEq \1( r( {x} ) = 0 ) \cdot q( {x}, y, h^{(1)}( {x} ) ) + \1( r( {x} ) = 1 ) \cdot q( {x}, y, h^{(2)}( {x} ) ) \\
    C( {x}, h^{\rm cas}( {x} ) ) &\defEq \1( r( {x} ) = 0 ) \cdot c^{(1)} + \1( r( {x} ) = 1 ) \cdot (c^{(1)} + c^{(2)}).
\end{align*}
Ideally, one seeks to maximize {quality} 
given a budget $B$ on average {inference cost}:
\begin{equation}
    \label{eqn:cascade-constraint}
    \max_{r \colon \XCal \to \{ 0, 1 \}} \mathbb{E}_{{x}, y}[ Q( {x}, y, h^{\rm cas}( {x} ) ) ] \colon \quad \mathbb{E}_{{x}}[ C( {x}, h^{\rm cas}( {x} ) ) ] \leq B.
\end{equation}
We note that the average cost $\mathbb{E}[ C( {x}, h^{\rm cas}( {x} ) ) ]$ is related to the \emph{deferral rate} $D( x ) = \Pr( r( x ) = 1 )$, via $\mathbb{E}[ C( {x}, h^{\rm cas}( {x} ) ) ] = c^{(1)} + D( x ) \cdot c^{(2)}$.
In practice, one may set $r( x ) = \1( s( x ) < t )$ for suitable $s \colon \XCal \to \Real$ and threshold $t \in \Real$.
One may choose $t$ to satisfy the inference cost constraint. 

Now, we discuss cascades for generative LMs. 
This largely follows the setup described above, except that we now consider \emph{probabilistic} models over \emph{sequences}.
Concretely, suppose we have two language models 
$p^{(1)}, p^{(2)}$,
with inference costs $c^{(1)}, c^{(2)}$.
Similarly,
suppose $q \colon \mathscr{V}^m \times \mathscr{V}^{m'} \times \Delta( \mathscr{V}^n ) \to \Real_+$ 
is a measure of the \emph{quality} of a given distribution over responses for a given prompt.
A cascade 
$p^{\rm cas}$ of these models
is parameterized by
a
deferral rule $r \colon \mathscr{V}^m \to \{ 0, 1 \}$,
and is given by:
$$ p^{\rm cas}( \cdot \mid \boldsymbol{x} ) = 
\1( r( \boldsymbol{x} ) = 0 ) \cdot p^{(1)}( \cdot \mid \boldsymbol{x} ) + \1( r( \boldsymbol{x} ) = 1 ) \cdot p^{(2)}( \cdot \mid \boldsymbol{x} ). $$
Given an input sequence $\boldsymbol{x}$ and target sequence $\boldsymbol{y}$,
an LM cascade results in quality and cost
\begin{align*}
    Q( \boldsymbol{x}, \boldsymbol{y}, p^{\rm cas}( \cdot \mid \boldsymbol{x} ) ) &\defEq \1( r( \boldsymbol{x} ) = 0 ) \cdot q( \boldsymbol{x}, \boldsymbol{y}, p^{(1)}( \cdot \mid \boldsymbol{x} ) ) + \1( r( \boldsymbol{x} ) = 1 ) \cdot q( \boldsymbol{x}, \boldsymbol{y}, p^{(2)}( \cdot \mid \boldsymbol{x} ) ) \\
    C( \boldsymbol{x}, p^{\rm cas}( \cdot \mid \boldsymbol{x} ) ) &\defEq \1( r( \boldsymbol{x} ) = 0 ) \cdot c^{(1)} + \1( r( \boldsymbol{x} ) = 1 ) \cdot (c^{(1)} + c^{(2)}).
\end{align*}
With these, we may construct a similar constrained objective as in Equation~\ref{eqn:cascade-constraint}.
{Similarly to the classification case, we may parameterize the deferral rule as $r(\boldsymbol{x}) = \1(s(\boldsymbol{x}) < t))$ for a suitable \emph{deferral score function} $s\colon \mathscr{V}^m \to  \mathbb{R}$ (which may depend on the output of $p^{(1)}( \cdot \mid \boldsymbol{x})$). 
We will investigate and analyze different types of deferral score functions on different NLP tasks.}

Recently,~\cite{Chen:2023a} introduced the FrugalGPT system to achieve efficient inference via multiple strategies, including LM cascades.
They also learn a deferral score to determine whether or not to terminate prediction;
however, this
depends on the input prompt and the generated output \emph{text},
and does not consider the model's token-level uncertainty as we shall explore subsequently. A few works have proposed to learn a router which can decide which model to use amongst a set of models depending upon the input prompt~ \citep{shnitzer2023large, hari2023tryage}. However, their settings do not necessarily consider models of increasing capacity and hence, their routers depend only on the input prompt not on the model confidence. 

%
\section{Confidence Measures for Language Model Cascades}
\label{sec:confidence_measures}
A key question in the design of cascades is the choice of deferral rule. 
{In this work, we seek to understand the behaviors of different types of deferral functions on NLP tasks. We start by discussing a few natural extensions of commonly used deferral rules for classification.}

\subsection{Chow-Sum and Chow-Average}

\textbf{Chow-Sum}. {We start with the multi-class classification setting where the output space $\mathscr{Y} = \{1, \ldots, L\}$ and $L \in \mathbb{N}_+$.}
In the simplest case, one may defer
if the \emph{confidence} in the prediction $h^{(1)}( x )$ of the small model is sufficiently low.
There are several means of quantifying confidence in classification~\citep{Shafer:2008,Guo:2017,Kendall:2017,Jiang:2018},
but arguably the simplest is
the \emph{predicted class probability}~\citep{Huang:2018,wang2022wisdom,Jitkrittum:2023},
which aligns with \emph{Chow's rule} from the closely related problem (see~\citet{Mozannar:2020,narasimhan2022posthoc}) of learning to reject~\citep{Cho1970}:
\begin{equation}
    \label{eqn:max-softmax}
    s( x ) \defEq p^{(1)}( \hat{y} \mid x ),
\end{equation}
where $p^{(1)}( \cdot \mid x )$ denotes the predictive distribution over possible labels of the small model,
and $\hat{y} = \argmax_{y \in \YCal} p^{(1)}( {y} \mid x )$ denotes the predicted label.

To design a deferral rule for LM cascading,
a natural starting point is to mimic the predicted class probability (Equation~\ref{eqn:max-softmax}):
we may compute
the (log) probability of the model generated sequence 
$\hat{\boldsymbol{y}}$,
\begin{align}
    \label{eqn:max-softmax-seq}
    s_{\rm sum}( \boldsymbol{x} ) &\defEq 
    \log  p^{(1)}( \hat{\boldsymbol{y}} \mid \boldsymbol{x} ) \\
    \label{eqn:max-softmax-seq-log}
    &= \sum_{i = 0}^{| \hat{\boldsymbol{y}} | - 1} \log p^{(1)}( \hat{y}'_{i + 1} \mid \boldsymbol{x}, \hat{y}'_{1}, \ldots, \hat{y}'_{i} ),
\end{align}
We term this approach \emph{{\tt Chow-Sum}}, as it involves the sum of per-token log probabilities.
Analogous to the prediction rule for classification,
we may set 
$\hat{\boldsymbol{y}} \defEq \argmax_{\boldsymbol{y} \in \mathscr{V}^*} p^{(1)}( \boldsymbol{y} \mid \boldsymbol{x} )$,
denoting by $\mathscr{V}^*$ the set of all sequences.
This requires searching over an exponentially large set;
however, efficient approximations via greedy or beam search are feasible.

\textbf{Chow-Average}.
{\tt Chow-Sum} computes the aggregate sequence-level log-probability. 
A natural variant is the
average of the per-token log-probabilities.
This is equivalently
the \emph{length normalized} log-probability,
or
the log-perplexity~\citep{Chen:1998}: 
\begin{equation}
    \label{eqn:max-softmax-seq-log-norm}
    s_{\rm avg}( \boldsymbol{x} ) \defEq \frac{1}{| \hat{\boldsymbol{y}} |} \sum_{i = 0}^{| \hat{\boldsymbol{y}} | - 1} \log p^{(1)}( \hat{y}'_{i + 1} \mid \boldsymbol{x}, \hat{y}'_{1}, \ldots, \hat{y}'_{i} ).
\end{equation}
Note that
$\hat{\boldsymbol{y}}$ may be computed as above,
without incorporating any length-normalization.

%
\subsection{Limitations of Chow-Sum and Chow-Average}
\label{sec:chow-fail}

Given that {\tt Chow-Sum} tracks closely with the well-established Equation~\ref{eqn:max-softmax}, 
it is tempting to conclude that this emphatically solves the LM cascade problem.
However, LMs can be susceptible to the \emph{length bias} problem~\citep{Murray:2018,Adiwardana:2020}:
shorter, lower quality responses may receive a higher probability than longer, higher quality responses.
This may be seen as a consequence of the fact that each 
$p( \cdot \mid \boldsymbol{x}, y_{1}, \ldots, y_{i} )$ provides an \emph{imperfect} estimate of a ``true'' probability $p^{*}( \cdot \mid \boldsymbol{x}, y_{1}, \ldots, y_{i} )$,
and that errors in these estimates compound with sequence length. 

The length-bias issue naturally suggests using an average instead of sum of log-probabilities.
However, {\tt Chow-Average} can over-correct for this length bias, 
and preferentially defer \emph{shorter} predictions. 
We will see concrete examples of this behavior in \S\ref{sec:chow-sum-bad}.
%
%

More fundamentally, both approaches 
are inherently limited in the way they aggregate token-level uncertainty.
In particular, computing the 
sum or average of per-token probabilities
may mask settings where \emph{individual tokens} are highly uncertain,
even if the entire sequence has reasonably high probability.
Such token-level uncertainty may be highly important in certain settings such as fact-answering:
here, an LM may be (correctly) highly confident on articles and other grammatical primitives, but these are of less interest than confidence on tokens corresponding to entities (say).
This observation has been previously noted and exploited to allow certain ``easy'' tokens to be quickly decoded~\citep{Schuster:2022}. This observation has also been exploited in knowledge distillation by using different teaching modes for ``easy'' versus ``hard'' tokens \citep{zhong2024revisiting}.

\begin{figure}[!t]
    \begin{minipage}[t]{0.49\linewidth}
    \vspace{0pt}
    \resizebox{0.99\linewidth}{!}{%
        \includegraphics[scale=0.25]{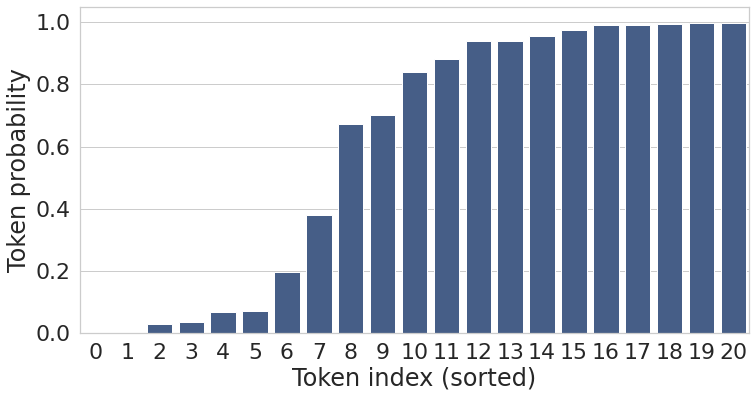}
    }
    \end{minipage}
    \begin{minipage}[t]{0.49\linewidth}
    \vspace{0pt}
    {\bf Model output tokens:}
    \begin{tcolorbox}
    {\tt 
The\_$\overbrace{\text{{\color{red}journey}}}^{1}$\_always\_$\overbrace{\text{{\color{red}leads}}}^2$\_to\_Es\_it\_j\_en\_i\_,
\_where\_the\_$\underbrace{\text{{\color{red}singer}}}_{3}$\_s\_visit\_their\_$\underbrace{\text{{\color{red}sponsor}}}_0$\_
children\_.
    }
    \end{tcolorbox} 
    \end{minipage}
    
    \caption{Example of tokenized FLAN-T5 Base model output on WMT FR $\to$ EN. 
    {\color{red}{Red}} tokens have a significantly higher uncertainty compared to the others, as shown in the left plot.
    (For each {\color{red}{red}} token, we note the rank of its uncertainty score in the right plot.)
    However, due to the large number of other more predictable tokens, {\tt Chow-Sum} gives the output a relatively high score. 
    }
    \label{fig:wmt-example-per-token}
\end{figure}

Figure~\ref{fig:wmt-example-per-token} presents an example of this phenomenon on the WMT FR $\to$ EN dataset (details in~\S\ref{sec:chow-generalisations}):
there can be cases where most tokens are highly predictable 
(i.e., $p( y'_i \mid x, y'_1, \ldots, y'_{i - 1} ) \approx 1$),
but a few tokens are less predictable
(i.e., $p( y'_i \mid x, y'_1, \ldots, y'_{i - 1} ) \approx 0$).
In such cases, {\tt Chow-Sum} can yield overly optimistic uncertainty estimates.
This motivates us to consider richer representations of uncertainty which can capture token-level uncertainty, instead of simply computing the sum or average over the sequence.

%
\subsection{Beyond Chow-Sum and Chow-Average: Chow-Quantile}

The discussion in \S\ref{sec:chow-fail} suggests there is value
in considering the following generalization of the maximal sequence probability:
\begin{equation*}
    s_{\rm quant}( \boldsymbol{x}, \alpha) \defEq {\tt quantile}_\alpha( p^{(1)}( \hat{y}'_1 \mid \boldsymbol{x} ), p( \hat{y}'_2 \mid \boldsymbol{x}, \hat{y}'_1 ), \ldots, p( \hat{y}'_{n} \mid \boldsymbol{x}, \hat{y}'_1, \ldots, \hat{y}'_{n - 1} ) ),
\end{equation*}
{where $\hat{\boldsymbol{y}} \defEq \argmax_{\boldsymbol{y} \in \mathscr{V}^*} p^{(1)}( \boldsymbol{y} \mid \boldsymbol{x} )$  is the most probable output sequence (or a suitable approximation thereof) under $p^{(1)}$.}
Here, ${\tt quantile}_\alpha$ computes the $\alpha$-quantile of the set of per-token log probabilities. For instance, $\alpha=0$ would correspond to taking the minimum per-token log probability as the deferral score.
One may regard quantiles as another way of converting the token-level uncertainty distribution into a single score, which are capable of capturing richer information from the token-level uncertainty distribution. For example, employing the \emph{maximal} token uncertainty 
(i.e., the \emph{minimum} of the per-token probabilities {\color{blue}{\citep{stengel2022calibrated}}})
can be useful in scenarios where most tokens are predictable, but a few important tokens are not (per Figure~\ref{fig:wmt-example-per-token}).

Next, we evaluate all the aforementioned approaches on multiple NLP tasks. For that, we describe the experimental setup used for the evaluation.

\subsection{Experimental Setup}


\textbf{Models}.
We employ
FLAN-T5~\citep{Chung:2022} models, which are T5 models~\citep{Raffel:2020} that have undergone instruction tuning~\citep{Wei:2022}.
This family offers a range of models of different sizes, 
spanning Small ($80$M parameters) to XXL ($11$B parameters),
and have demonstrated strong few-shot performance on a range of NLP benchmarks.
In the body, we primarily focus on a two-model cascade of FLAN-T5 Base and FLAN-T5 Large. Results for other models are included in the Appendix.
We employ these models with few-shot prompting and greedy decoding.

%
\textbf{Evaluation}.
We summarize performance using the \emph{deferral curve}.
Consider a candidate deferral rule produced by thresholding 
$s( \boldsymbol{x} ) \in \Real$
via $r( \boldsymbol{x} ) = \1( s( \boldsymbol{x} ) < t )$.
Let $p^{\rm cas}$ denote the associated cascaded LM.
For a fixed threshold $t$, we may compute the associated
\emph{deferral rate} $\Pr( r( \boldsymbol{x}) = 1 )$,
and the associated \emph{cascade quality} $\mathbb{E}[ Q( \boldsymbol{x}, \boldsymbol{y}, p^{\rm cas}( \cdot \mid \boldsymbol{x} ) ) ]$.
The deferral curve is produced by
plotting the trade-off between deferral rate and cascade quality
as $t$ is varied.
As a scalar summary,
we report
the \emph{area under the deferral curve} (AUC-DF).
For a given dataset, higher AUC-DF values indicate better deferral curves.
Note that the range of AUC-DF values vary across datasets, however.

%
\textbf{Datasets}.
In the body, we show deferral curves for three different NLP tasks:
MNLI~\citep{Williams:2018}, a multi-class classification problem;
TriviaQA~\citep{Joshi:2017}, a closed-book question answering problem;
and
WMT DE $\to$ FR, a translation problem.
We report AUC-DF numbers for an expanded dataset pool.
These span
\emph{Classification}
(IMDb~\citep{Maas:2011},
SuperGLUE~\citep{Wang:2019b},
MNLI~\citep{Williams:2018},
ANLI~\citep{Nie:2020});
\emph{Question answering} 
(TriviaQA~\citep{Joshi:2017},
NaturalQA~\citep{Kwiatkowski:2019},
TyDiQA \{ ID, SW, FI \}~\citep{Clark:2020});
\emph{Reading comprehension}
(Lambada~\citep{Paperno:2016},
SQuAD~\citep{Rajpurkar:2016});
\emph{Translation}
(WMT 14: EN $\to$ FR \citep{bojar-EtAl:2014:W14-33}, WMT 19: DE $\to$ FR \citep{wmt19translate}, and WMT 14: FR $\to$ EN \citep{bojar-EtAl:2014:W14-33});
and
\emph{Common-sense reasoning}
(Winogrande~\citep{Sakaguchi:2020}). Note that we treat all problems as finding a text to text mapping. So for classification tasks, we encode the classes as strings. For evaluation, we take the model’s output text and perform a string comparison to the label. See Table~\ref{tbl:data-info} (Appendix) for more details.
\subsection{Evaluating Confidence Measures for Cascades}
\label{sec:chow-generalisations}

We now empirically validate the critiques in \S\ref{sec:chow-fail}, demonstrating that using the standard sequence probability ({\tt Chow-Sum}) to defer can result in overly penalizing longer sequences. Moreover, {\tt Chow-Average} flips this bias and overly defers shorter sequences.
We then verify that the {\tt Chow-Quantile} generalization proposed above can capture richer token-level uncertainty.

\begin{figure*}[!t]

    \resizebox{0.99\linewidth}{!}{%
        \includegraphics[scale=0.3]{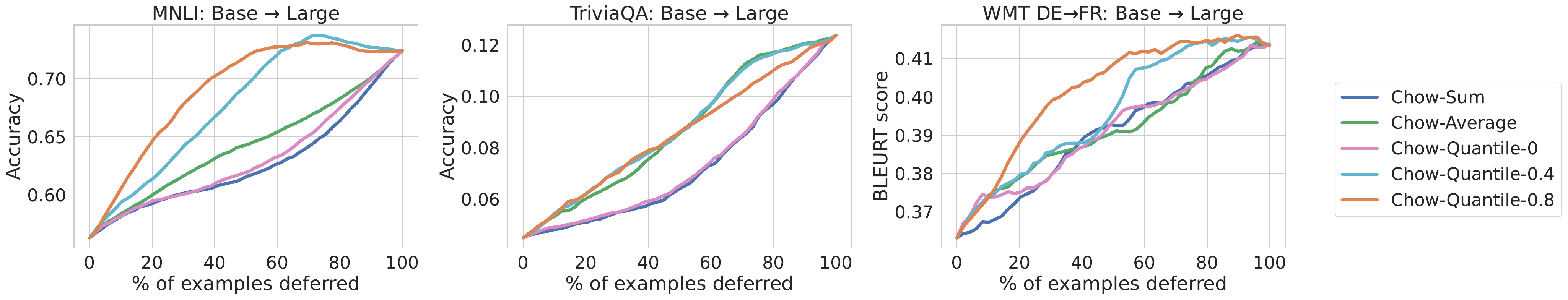}
    }
    
    \caption{Deferral curves on MNLI, TriviaQA, and WMT DE $\to$ FR for a FLAN-T5 Base $\to$ Large cascade.
    {\tt Chow-Quantile} consistently outperforms {\tt Chow-Sum} and {\tt Chow-Average}.
    This confirms there is value in going beyond na\"{i}ve sequence probability as an uncertainty measure for cascading.}
    \label{fig:chow_sum}
\end{figure*}

\begin{figure*}[!t]
    \centering

    \hfill    
    \begin{subfigure}{0.28\textwidth}
    \centering
     \includegraphics[width=\linewidth]{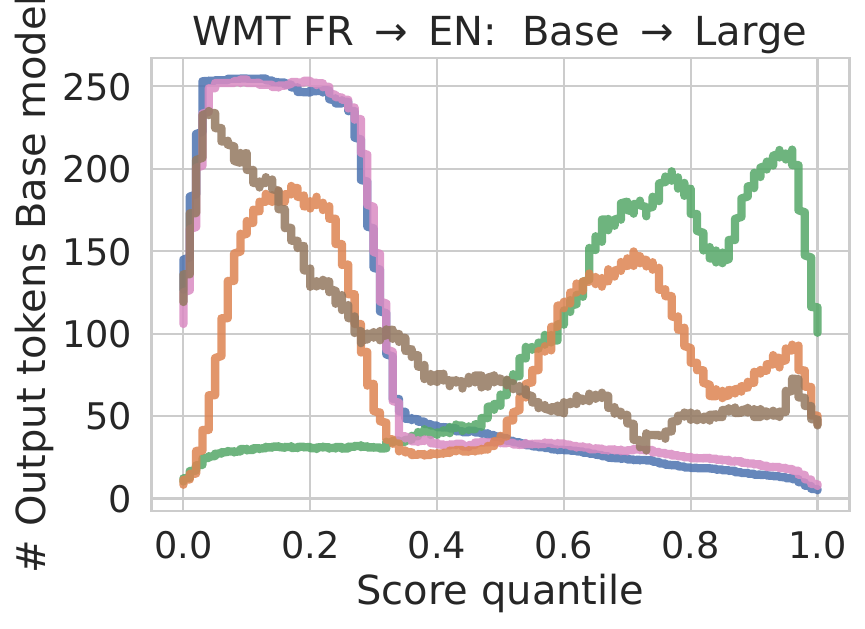}
    \caption{}
    \label{fig:wmt-length-plot}
    \end{subfigure}
    \hfill
    \begin{subfigure}{0.28\textwidth}
    \centering
     \includegraphics[width=\linewidth]{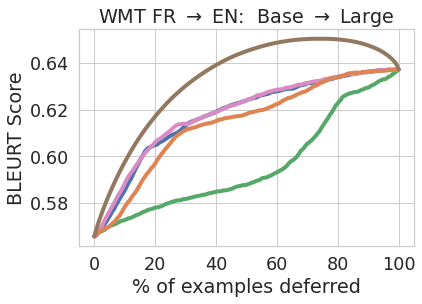}
     \caption{}
    \label{fig:wmt-cascade-curve}
    \end{subfigure}
    \begin{subfigure}{0.12\textwidth}
    \centering
     \includegraphics[width=\linewidth]{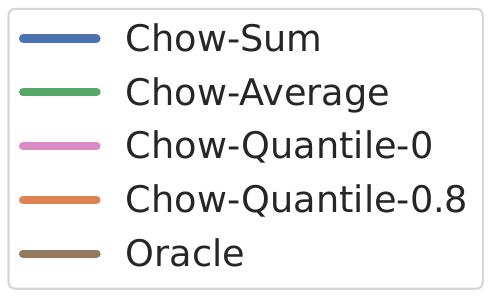}
    \end{subfigure}
     \hfill
    \begin{subfigure}{0.28\textwidth}
    \centering
     \includegraphics[width=\linewidth]{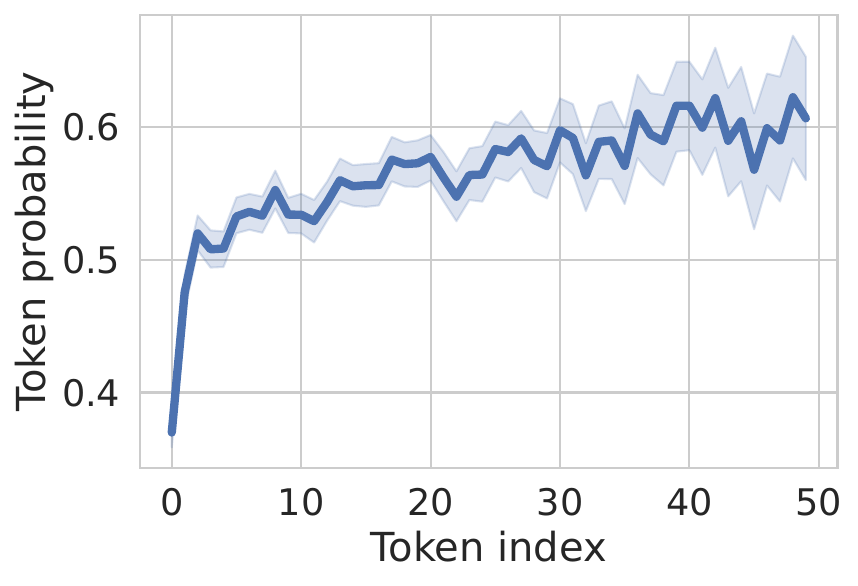}    
     \caption{}
    \label{fig:decreasing_probability}
    \end{subfigure}
     \caption{
{\bf (a)}
Relation between deferral rules and output length (number of tokens) for WMT FR $\rightarrow$ EN dataset and FLAN-T5 Base Model. 
{\tt Chow-Sum} tends to defer longer prompts:
the prompts with lowest scores have notably higher length than those with higher scores.
Interestingly, {\tt Chow-Average} \emph{over-}corrects this bias:
it tends to overly defer prompts with \emph{lower} length. {\tt Chow-Quantile-0} again defers longer outputs more whereas {\tt Chow-Quantile-0.8} initially focuses more on the shorter outputs.  Oracle refers to deferring using the difference of BLEURT scores of the two models. 
Oracle also tends to defer longer outputs, but the preference is moderate as compared to {\tt Chow-Sum}. 
{\bf (b)} Corresponding deferral curves. 
%
{\bf (c)} Analysis of token-level uncertainty on WMT FR $\to$ EN.
For each token index $i$,
the 
corresponding average
prediction probability across all examples (with prediction length $\leq i$) 
for FLAN-T5 Base. 
We observe that later tokens tend to have higher probability, i.e.,
the model is generally the most uncertain for early tokens.
}
    \vspace{-5mm}
    \label{fig:wmt_length_analysis}
\end{figure*}

%
\textbf{Summary of results}.
Figure~\ref{fig:chow_sum} plots the deferral curves for three datasets. We see that 
(for a particular choice of quantile),
{\tt Chow-Quantile} consistently outperforms standard {\tt Chow-Sum} and {\tt Chow-Average}. 
AUC-DF values for all datasets are included in Table~\ref{tbl:area_under_curve}. Looking at the table, we see that while a \emph{particular} choice of quantile is able to do well,
there is no single consistent choice that performs well across tasks. 
Next, we discuss insights into the results by using WMT FR$\rightarrow$EN as an example dataset.




\textbf{Why can Chow-Sum and Chow-Average be sub-optimal?}
\label{sec:chow-sum-bad}
To better understand the reason for the sub-optimality of {\tt Chow-Sum}, Figure~\ref{fig:wmt_length_analysis} studies the relation between the deferral rule and output length.
Specifically, 
for each test prompt $\boldsymbol{x}$,
let $\hat{\boldsymbol{y}}$ denote the result of decoding via the small model in the cascade.
For each deferral rule,
we compute the corresponding score $s( \boldsymbol{x} )$ and the length $| \hat{\boldsymbol{y}} |$.
For ease of comparison across different rules, we convert each score into the corresponding quantile.

Figure~\ref{fig:wmt_length_analysis} reveals that {\tt Chow-Sum} tends to defer prompts with larger output lengths:
the prompts with lowest scores have notably higher output length than those with higher scores. This makes us  ask if it is all bad to defer prompts with longer outputs? 
We observe that for the Base model on the WMT datasets, 
even the BLEURT \citep{sellam2020bleurt} scores 
tend to have a non-zero negative correlation with output lengths
(Table~\ref{tbl:wmt_bleurt_length_correlation}, Appendix). 
A closer look at the model predictions shows that longer predictions tend to have repetitions, as shown in Figure~\ref{fig:predictions-examples} (Top) and hence, are good candidates for deferral. 
(The shown predictions are truncated for clarity.)

\begin{figure}[!t]

    \small
    \begin{tabular}{@{}p{2.625in}p{2.625in}@{}}
    \toprule
    {\tt For the Australian Government, Keith Brown called Mr. Carmichael "unjustly" to support the inclusion of the Ecosse in the HS2.net network. ?? ?? ?? ?? ?? ?? ?? ?? ?? ....
    } &
    {\tt The announcement of the release of a new album by David Bowie has left everyone a little bit a little a little a little a little a little bit a little bit a little bit a little bit ....
    } \\
    \midrule
    {\tt The lyric Adéro, a concert-recording held last August, was added to the repertoire, with the competition of the coloratura soprano Marie- ?? ve Munger. The lyric Adéro was added to the repertoire in August. The lyric Adéro was added to the repertoire in August. ....
    } &
    {\tt "The Emergency Room at the Hotel-Dieu must be closed as soon as possible, and for us, it is November 4th," says Lo ?? c Capron, Chairperson of the APHP Medical Committee (MCM), who supports the direction.
    } \\
    \midrule
    {\tt Obama lays out his reform plan
    } &
    {\tt A memorial ceremony for the victims
    } \\
    \bottomrule
    \end{tabular}

    \caption{FLAN-T5 Base predictions on WMT FR $\to$ EN. {\bf Top:} Predictions with the longest lengths. These tend to have repetitions, indicating low quality output that could be resolved with a larger model; 
    length does have some signal in identification of good candidates for deferral. 
    {\bf Middle:} The predictions which {\tt Chow-Quantile-0} tends to defer. This quantile tends to identify repetitions and ``{\tt??}" (unknown tokens) as these tokens tend to have lower probability. 
    {\bf Bottom:} The predictions which {\tt Chow-Quantile-0.8} tends to defer. This quantile prioritizes deferring shorter inputs.}
    \label{fig:predictions-examples}
    \vspace{-5mm}  
\end{figure}    

%
%


%



This shows that there is some signal in output length as a deferral score.
However, {\tt Chow-Sum} is overly biased towards deferring longer predictions and hence, can be sub-optimal. Interestingly, {\tt Chow-Average} \emph{over-}corrects this bias:
it tends to overly defer prompts 
with \emph{lower} output length.

\textbf{Why does Chow-Quantile help?}
As discussed above, {\tt Chow-Quantile} is able to capture rich information from the token-level uncertainty vector. We discuss below why {\tt Chow-Quantile-0} and {\tt Chow-Quantile-0.8} work well with respect to the WMT FR$\rightarrow$EN dataset.


{\tt Chow-Quantile-0}: The main insight is that the minimum token probability 
is able to capture repetitions and ``{\tt ??}" (unknown tokens), 
as they generally tend to have lower probability values and are more uncertain.
This confirms our understanding that quantiles can capture richer token-level uncertainty. 
We show two examples with the minimum {\tt Chow-Quantile-0} value for the WMT FR$\rightarrow$EN dataset and FLAN-T5 Base in Figure~\ref{fig:predictions-examples} (Middle).



{\tt Chow-Quantile-0.8}: Interestingly, {\tt Chow-Quantile-0.8} tends to defer shorter predictions. We show two examples with the minimum {\tt Chow-Quantile-0.8} value in Figure~\ref{fig:predictions-examples} (Bottom). 
To understand this,
Figure~\ref{fig:decreasing_probability} shows the 
average token probability as a function of the token index,
for the
WMT EN $\rightarrow$ FR dataset and FLAN-T5 Base model. 
As the token index increases, the average probability increases; i.e., the model tends to become more confident. Hence, the {\tt Chow-Quantile-0.8} is able to focus more on the shorter, uncertain outputs.

In summary, we have seen that {\tt Chow-Quantile-0} is able to focus more on identifying the presence of repetitions and unknown tokens ``{\tt ??}'' whereas {\tt Chow-Quantile-0.8} is able to capture the uncertainty in shorter predictions better. Thus, we conclude that different quantiles are able to capture richer and complementary measures of uncertainty. Moreover, we have already seen that there is no one quantile which works well across all datasets. Given this, a natural option is to learn how to combine various quantiles for a given dataset, which we consider next.

\section{Post-hoc Deferral Rules}
\label{sec:post_hoc_routers}
We show that training 
\emph{post-hoc} deferral rules based on probability quantiles, 
and (optionally) suitable embeddings from the small and large model, 
can significantly improve the cost-quality tradeoff.

\subsection{Post-hoc deferral rule training}

The idea of learning when to defer in a cascade 
follows a recent line of work on  classification~\citep{narasimhan2022posthoc,Kag:2023,Jitkrittum:2023}.
In a nutshell, 
for suitable feature mapping $\Phi( \boldsymbol{x} ) \in \Real^d$,
we seek to learn a deferral score $s( \boldsymbol{x} ) \in \Real$ via a standard model class (e.g., a feedforward network).
We then defer using $r( \boldsymbol{x} ) = \1( s( \boldsymbol{x} ) < t )$.

To construct the input features, we set $\Phi( \boldsymbol{x} )$ to be a fixed length vector comprising the per-token probability quantiles from the small model.
Additionally, we add the aggregate scores from {\tt Chow-Sum} and {\tt Chow-Average} (see Appendix~\ref{phi-setup}).
To fit the deferral scorer on a training set of input prompts $\{ \boldsymbol{x}_i \}$, we minimize an empirical loss against a set of target labels $\{ z_i \}$.
For tasks based on accuracy, 
we set $z_i = 1$ iff the large model is correct, and the small model is incorrect on the given example; 
i.e., it would benefit to defer to the larger model. 
We then fit the scorer with the binary logistic loss.
For translation tasks,
the target is the difference of BLEURT scores of the two models; 
we train with the square loss. We call this method {\tt Post-Hoc-Quantile} (see Appendix~\ref{sec-exp-details} for details).


%
\subsection{Leveraging intermediate embeddings}

The above target labels exploit information from the large model during \emph{training}.
Importantly, we cannot directly use such information during \emph{inference}, as it would require querying the large model (and thus defeat the point of cascading).
Note, 
however, that in some settings it may be feasible to use
\emph{intermediate} information from the large model, 
e.g., 
token embeddings from an intermediate layer.
Prior work has noted that such intermediate embeddings can often contain valuable information by themselves~\citep{Schuster:2022}.

Inspired by this, we thus study the viability of using such intermediate embeddings for training post-hoc deferral rules.
For encoder-decoder models such as T5, 
such embeddings can be from either the encoder, decoder, or both. We study two methods - one which uses the final decoder embeddings of the smaller model averaged over all tokens. We call this method {\tt Post-Hoc-Embed-1}. In the second method, we  add the first token embedding from the first decoder layer of the large model as another input to the post-hoc rule. We call this method {\tt Post-Hoc-Embed-1+2}.

We remark that previous work~\citep{Ren:2023} has shown the value of
final layer embeddings for \emph{selective generation},
and the related problem of \emph{out-of-distribution} detection.
We caution also that while not as expensive as querying the entire large model,
even extracting intermediate decoder embeddings can involve a non-trivial cost.
Nonetheless, in settings where some increase in cost is acceptable, it is of interest whether these embeddings offer significantly valuable information.

%

%
\subsection{How well does post-hoc deferral work?}
\label{sec:router_experiments}

%

\begin{figure*}[!t]

    \resizebox{0.99\linewidth}{!}{%
          \includegraphics[scale=0.3]{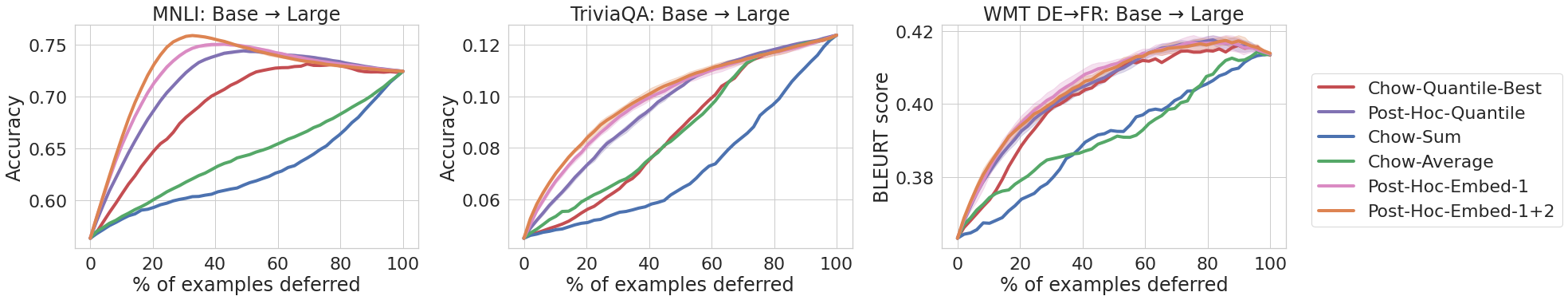}
    }
    
    \caption{Deferral curves on MNLI, TriviaQA, and WMT DE $\to$ FR for a FLAN-T5 Base $\to$ Large cascade.
    The post-hoc deferral rule yields consistently on-par or superior performance compared to other methods.
    Further exploiting intermediate embeddings of the large model yields large gains.
    See Table~\ref{tbl:area_under_curve} for a summary of deferral curves on multiple datasets. 
    }
    \vspace{-2mm}
    \label{fig:post_hoc_deferral rules}
\end{figure*}

%
For the same experimental setup as the previous section,
Table~\ref{tbl:area_under_curve} summarizes the {area under the deferral curve} (AUC-DF) across various datasets. 
Numbers are averaged over 5 random runs.
We see that the post-hoc deferral rule approach consistently performs the best;
in scenarios where other methods are better, the difference is minimal.
For a more fine-grained analysis, Figure~\ref{fig:post_hoc_deferral rules} plots the full deferral curves on MNLI, TriviaQA, and WMT. In the plot, {\tt Chow-Quantile-Best} chooses the best method out of {\tt Chow-Quantile-*} based on the validation data.
We see that post-hoc routing is generally on-par with the {\tt Chow-*} family of (non-learned) deferral rules.

We see that {\tt Post-Hoc-Embed-1} is able to improve upon {\tt Post-Hoc-Quantile} and {\tt Chow-Quantile-*} methods slightly but is slightly inferior compared to the {\tt Post-Hoc-Embed-1+2} method. This intuitively makes sense as this has more information compared to the {\tt Post-Hoc-Quantile} method but still does not include any information about model 2.

Strikingly, further using the larger model's intermediate embeddings can lead to significant improvements across all deferral rates,
particularly for \emph{classification tasks}.
Intuitively, the first token's intermediate embedding could have a lot of information for classification tasks, where the answers typically comprise of a single word and only a couple of tokens. 
However, for generation and translation tasks with longer outputs, the main token containing the answer could be present later in the sequence and thus, there may be limited use of using the first token embedding. One may wonder if we really need quantiles to train the post-hoc deferral rule. 
In Appendix~\ref{appdx:feature_ablations}, we show that na\"{i}vely passing the probability vector with padded zeros performs poorly in many cases. 

%
\begin{table}[!t]
    \centering
    
    \resizebox{\linewidth}{!}{
\begin{tabular}{@{}lccccccc@{}}
\toprule
\toprule
   {\bf Dataset} &    {\bf\tt Chow-Sum} & {\tt Chow-Average} & {\bf\tt Chow-Quantile-0} & {\bf\tt Chow-Quantile-0.4} & {\bf\tt Chow-Quantile-0.8} &       {\bf\tt Post-Hoc-Quantile}&     {\bf\tt Post-Hoc-Embed-1+2} \\
\midrule
   ANLI-R1 &        0.524 (+4.59)  &        0.519 (+3.59) &    \bfseries{0.534 (+6.58)} &             0.515 (+2.79) &             0.515 (+2.79) & \bfseries{0.534 (+6.58)} & \bfseries{\textcolor{mydarkblue}{0.563 (+12.51)}} \\
   ANLI-R2 &        0.446 (+0.67) &        0.446 (+0.67) &    \bfseries{0.449 (+1.35)} &             0.440 (-0.67) &             0.441 (-0.45) &        0.442 (-0.22) & \bfseries{\textcolor{mydarkblue}{0.462 (+4.36)}} \\
   ANLI-R3 &        0.413 (+3.50) &        0.422 (+5.76) &           0.417 (+4.51) &      \bfseries{0.425 (+6.51)} &             0.424 (+6.26) & \bfseries{0.425 (+6.51)} & \bfseries{\textcolor{mydarkblue}{0.440 (+10.45)}}\\
     BoolQ  &        0.838 (+2.57) &        0.838 (+2.57) &           0.838 (+2.57) &             0.838 (+2.57) &             0.838 (+2.57) &        \bfseries{0.840 (+2.81)} & \bfseries{\textcolor{mydarkblue}{0.854 (+4.64)}}\\
      IMDb  & \bfseries{0.964 (+1.15)} & \bfseries{0.964 (+1.15)} &    \bfseries{0.964 (+1.15)} &      \bfseries{0.964 (+1.15)} & \bfseries{0.964 (+1.15)} & \bfseries{0.964 (+1.15)} & \bfseries{\textcolor{mydarkblue}{0.965 (+1.31)}} \\
   Lambada  & \bfseries{0.703 (+3.38)} &        0.702 (+3.23) &    \bfseries{0.703 (+3.38)} &             0.694 (+2.05) &             0.687 (+1.02) &        0.701 (+3.08) & 0.692 (+1.82) \\
      MNLI  &        0.627 (-2.63) &        0.642 (-0.31) &           0.631 (-2.01) &             0.677 (+5.12) &             0.691 (+7.29) & \bfseries{0.711 (+10.4)} &\bfseries{\textcolor{mydarkblue}{ 0.722 (+12.24)}} \\
      PIQA  &        0.712 (+0.99) &        0.708 (+0.42)  &           0.715 (1.41) &             0.706 (+0.14) &             0.705 (+0.00) & \bfseries{0.717 (+1.70)} & \bfseries{\textcolor{mydarkblue}{0.727 (+3.26)}} \\
     SQuaD  &        0.410 (+2.24) &        0.408 (+1.74) &           \bfseries{0.410 (+2.24)} &             0.399 (-0.49) &             0.398 (-0.74) &        0.409 (+1.99) & \bfseries{\textcolor{mydarkblue}{0.412 (+2.76)}} \\
  TriviaQA  &        0.073 (-13.09) &        0.087 (+3.57) &           0.073 (-13.09) &             0.088 (+4.76) &             0.086 (+2.38) & \bfseries{0.097 (+15.47)} & \bfseries{\textcolor{mydarkblue}{0.100 (+19.56)}} \\
 TyDiQA-FI & \bfseries{0.332 (+3.10)} &        0.322 (+0.00) &           0.328 (+1.86) &             0.318 (-1.24) &             0.312 (-3.10) &        0.331 (+2.79) & \bfseries{\textcolor{mydarkblue}{0.338 (+5.27)}}\\
 TyDiQA-ID & \bfseries{0.247 (+7.39)} &        0.239 (+3.91) &           0.245 (+6.52) &             0.230 (+0.00) &             0.230 (+0.00) &        0.235 (+2.17) & 0.242 (+5.35) \\
 TyDiQA-SW &        \bfseries{0.170 (+9.67)} &        0.147 (-5.16) &           0.166 (+7.09) &             0.141 (-9.03) &             0.140 (-9.67) &        0.163 (+5.16) & 0.162 (+4.98) \\
Winogrande &        0.648 (+2.36) &        0.648 (+2.36) &           0.648 (+2.36) &      \bfseries{0.649 (+2.52)} &             0.648 (+2.36) &        0.648 (+2.36) & \bfseries{\textcolor{mydarkblue}{0.670 (+5.87)}} \\
 WMT DE $\rightarrow$ FR &        0.390 (+0.00) &        0.392 (+0.51) &           0.391 (+0.25) &             0.396 (+1.53) &             0.401 (+2.82) & \bfseries{0.404 (+3.58)} & \bfseries{\textcolor{mydarkblue}{0.404 (+3.69)}} \\
 WMT EN $\rightarrow$ FR &        0.436 (+1.16) &        0.435 (+0.92) &           0.436 (+1.16) &             0.435 (+0.92) &             0.439 (+1.85) & \bfseries{0.441 (+2.32)} & \bfseries{\textcolor{mydarkblue}{0.444 (+3.05)}} \\
 WMT FR $\rightarrow$ EN &        0.618 (+2.82) &        0.595 (-0.99) &           0.618 (+2.82) &             0.595 (-0.99) &             0.614 (+2.16) & \bfseries{0.619 (+2.99)} & {0.618 (+2.89)} \\
\bottomrule
\end{tabular}
}
    \caption{Table showing area under the deferral curve (AUC-DF). Numbers in brackets indicate \% change over the random baseline.
    The post-hoc deferral rule approach is the most consistently performant method. We use bold \& black to denote the best method amongst {\tt Chow-*} and {\tt Post-Hoc-Quantile} methods. We color {\tt Post-Hoc-Embed-1+2} with blue if it is the best amongst all.
    }
    \label{tbl:area_under_curve}
\end{table}

%

\section{Discussion and Future Work}
We have seen that there is value in going beyond simple sequence level uncertainty and considering finer measures of token level uncertainty as deferral rules for cascades. Moreover, we have seen that 
intermediate embeddings from the larger model can further boost performance. 

This work raises a number of interesting directions for future work.
First, we have used a $5$-layer MLP to train post-hoc deferral rules using the quantiles of the probability distribution. It would be interesting to see how well a $1$-layer Transformer works for this task, to potentially better exploit the sequential nature of the token probability vector.
%
We have focused on FLAN-T5 instruction tuned Encoder-Decoder models in this work. We believe that the insights and methods should generalize to Decoder-only architectures. It would be interesting to evaluate the proposed approaches for such architectures. Moreover, it has been observed that models with RLHF finetuning become uncalibrated (Figure 8 in~\citet{OpenAI2023GPT4TR}). It would be interesting to see how various finetuning steps affect the findings in this work and what consequences they have for designing efficient cascades.

Multiple works have considered alternative notions of uncertainty using the generative abilities of LMs, for example, reprompting the model to ask how confident it is about the answer or output an additional confidence score as part of the outputs~\citep{Kadavath:2022}. 
It would be interesting to evaluate how well these measures work for cascades which we discuss in the next section.



\section{Uncertainty Quantification for LMs}

\label{sec:uncertainty-quant}
There has been a large body of work on uncertainty quantification for LMs. We discuss some of the approaches below. They can be broadly divided into the following categories.

\textbf{Consensus-based}.
One limitation of Equation~\ref{eqn:max-softmax-seq} 
is that it considers a \emph{single} output sequence, e.g., the most likely one.
However, as there are many sequences that have similar meaning, it is intuitively more reliable to consider drawing \emph{multiple} sequences from $p( \cdot \mid \boldsymbol{x} )$.
One may then assess the \emph{consensus} in resulting predictions to measure confidence~\citep{Wang:2023,Xiao:2021,Chen:2023b};
this can help distinguish between models that are locally diffused versus peaked around a candidate sequence. Recently,~\citet{Yue:2023} explored the use of answer consistency to construct effective cascades.

\textbf{Deep ensembles and dropout}.
One approach to measure confidence is to create an ensemble of different models, and suitably aggregate their predictions (e.g., based on disagreement)~\citep{Landeghem:2022,Wang:2019,Gleave:2022}. However, these uncertainty estimation procedures involve additional computation (e.g., multiple inferences with a single model in dropout-based approaches and single inference with multiple models in ensemble-based approaches) compared to simply using softmax probability outputs from a single network. Such approaches are less appealing for use in cascades, where the primary goal is to improve efficiency.

\textbf{Post-hoc calibration/Answer- and length-bias calibration}.
For tasks involving question-answering with multiple choices (e.g., {\tt (A)}, {\tt (B)}, {\tt (C)}),
several works have demonstrated that LMs can have prior biases to certain answers, which can be identified and corrected~\citep{Zhao:2021,Holtzman:2021,Kumar:2022,Murray:2018,Mielke:2022,Jiang:2021}.

\textbf{Semantic entropy}.
Another key challenge in measuring the uncertainty for natural language outputs is that there are a lot of semantic equivalent sentences and hence, the probability can be divided among multiple outputs which mean the exact same thing. \citet{Kuhn:2023} proposes to mitigate this problem by sampling multiple outputs and then clustering semantically equivalent outputs together and combining their probability together. It would be interesting to understand how well this method can work for our setting.  

\textbf{Generative uncertainty}.
The above has largely focussed on generalizing the standard maximum predictive probability (Equation~\ref{eqn:max-softmax}) from classification to the LM setting.
While this by itself leads to a rich array of possible confidence measures,
LMs intriguingly offer a wholly new possible means of assessing confidence:
one may directly probe the model to obtain how confident it is on the proposed answer ~\citep{Kadavath:2022}. \cite{Kadavath:2022} discuss various ways of the input prompt format for this confidence probe. 
They also discuss the training of an additional head of the model to predict the model confidence but again, it is not clear how this compares with the standard probability output by the model without any additional finetuning. However, \citep{shrivastava2023llamas} found that the confidence measures generated linguistically give worse estimates of uncertainty compared to the classical softmax-based measures even when these softmax-based probabilities come from a different and weaker model. Moreover, they observed that two sources of uncertainty are complementary and it can be beneficial to combine them.

%
\textbf{Other work}.
\citet{Zhao:2023} proposed sequence-level calibration as a means to improve the generative ability of LMs; such calibration could also be useful in improving methods such as {\tt Chow-Sum}.
\citet{Kuhn:2023b} proposed to ask the model to detect ambiguous questions which the model is likely to get wrong and answer clarifying questions if the question is indeed ambiguous. {\citet{Hendy:2023}} proposed to use an exogeneous quality estimation model to decide how to route between two models.
\citet{Sakota:2023} similarly proposed to train a meta-model to pick an appropriate model from a family. \cite{fadeeva2023lm} did a comprehensive experimental analysis of various uncertainty methods.


%

%

\newpage
\bibliography{references}
\bibliographystyle{iclr2024_conference}


\clearpage
\newpage


\begin{center}
{\LARGE{}\ourtitle{}}{\LARGE\par}
\par\end{center}


\appendix


\addcontentsline{toc}{section}{Appendix} 
\part{Appendix} 
\parttoc 

%
\section{Experimental details}
\label{sec-exp-details}
We use a Multi-Layer-Perceptron (MLP) with 5 layers and 32 hidden units per dimension for training post-hoc deferral rules with Quantiles with batch normalization layers. We train for 200 epochs and early stop depending on the AUC of predicting the target label on the validation set for classification tasks and validation regression loss for the regression tasks. Since the embeddings are high dimensional, we use MLP with 2 layers and 8 hidden units to prevent overfitting. We train for the same number of epochs with early stopping. We use ADAM optimizer with a learning rate of $10^{-5}$. We use a batch size of 16 for training. We train over 5 random runs and show mean and standard deviations over the runs. 

We limit the number of examples per dataset per split to be 10k. Since many datasets do not have all the three splits available. We split the training set into train (0.8 fraction) and validation sets (0.2 fraction). And, use the validation or test split for reporting numbers. The details of all the datasets are included in Table~\ref{tbl:data-info}.

\begin{table}[!h]
\begin{center}
\begin{tabularx}{\textwidth}{@{}cccXXXXXX@{}} 
 \hline
 \multirow{2}{*}{Dataset} & \multirow{2}{*}{Category} &
 \multirow{2}{*}{\# Shots} &
 \multicolumn{2}{c|}{Train} & \multicolumn{2}{c|}{Test} & \multicolumn{2}{c|}{\# target tokens} \\
 \cline{4-9}
  &  &  & Split & Size & Split & Size & Mean & Std \\
 [0.5ex] 
 \hline
 BoolQ & Classification & 2 & Train & 9417 & Val & 3270 & 1 & 0 \\ 
 IMDb & Classification & 2 & Train & 9991 & Test & 9981 & 1 & 0 \\
 MNLI & Classification & 3 & Train & 10000 & Test & 9832 &  1.9 & 1.4\\
 ANLI-R1 & Classification & 3 & Train & 9999 & Test & 1000  & 1.9 & 1.3\\
 ANLI-R2 & Classification & 3 & Train &  10000 & Test & 1000  & 1.9 & 1.4\\
 ANLI-R3 & Classification & 3 & Train & 10000 & Test & 1200 & 1.9 & 1.3 \\ 
 PIQA & Classification & 2 & Train & 10000 & Val & 1838 & 24.4 & 25.3\\
 Winogrande & Classification & 2 & Train & 9998 & Val & 1267 & 1.3 & 0.7 \\
 TriviaQA & Generation & 2 & Train & 10000 & Val & 10000 & 4.0 & 3.3\\
 NaturalQA & Generation & 2 & Train & 10000 & Val & 3610 & 3.5 & 2.1 \\ 
 SQuaD & Generation & 2 & Train & 10000 & Val & 9997 & 4.6  & 4.1 \\
 Lambada & Generation & 2 & Train & 4857 & Test & 5152 & 2.0 & 1.1 \\ 
 TyDiQA-FI & Generation & 2 & Train & 3659 & Val & 782 & 14.1 & 27.4 \\
 TyDiQA-SW & Generation & 2 & Train & 3611 & Val & 499 & 12.2 & 24.3 \\
 TyDiQA-ID & Generation & 2 & Train & 3655 & Val & 565 & 10.3 & 21.0\\
 WMT FR$\rightarrow$EN & Translation & 2 & Train  & 9976 & Test & 3003 &  35.3 & 40.7 \\ 
 WMT EN$\rightarrow$FR & Translation & 2 & Train & 9974 & Test & 3003 & 54.3 & 48.5 \\
 WMT DE$\rightarrow$FR & Translation & 2 & Train & 9999 & Val & 1497 &  33.6 & 23.2 \\
 \hline
\end{tabularx}
\end{center}
\caption{The table shows the number of shots used for each dataset, the splits used for training and testing, the number of examples in each split and the average number of target tokens in the train split of each of the datasets.}
\label{tbl:data-info}
\end{table}

%

%

%

%
\clearpage
\section{Additional experimental results}
\subsection{Ablations with probability vectors as features}
\label{appdx:feature_ablations}

In this section, we show that simply using the sequence of probability values with padded zeros to account for variable length sequences does not work as well as using quantiles. This intuitively makes sense, as the different token uncertainties are not aligned with each other which makes learning hard for the MLP. We further show that using sorted probability values does not work as well as using the quantiles: quantiles help better align the different uncertainty scores for variable length sequences.

\begin{figure}[H]
    \centering
    \includegraphics[scale=0.18]{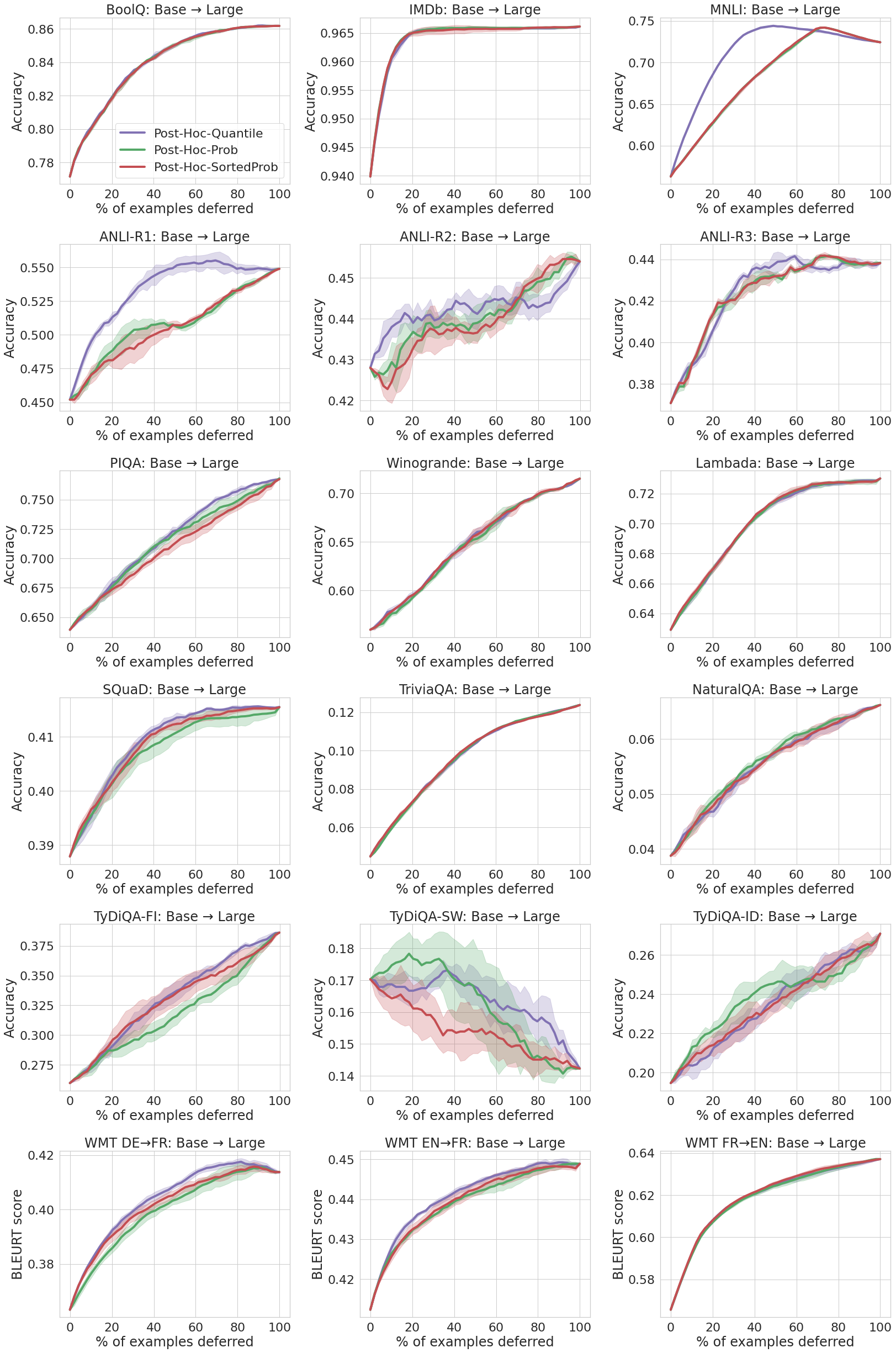}
    \caption{Deferral curves on different datasets for a FLAN-T5 Base $\to$ Large cascade.
    Post-Hoc-Quantile denotes the post-hoc deferral rule which includes quantile features. 
    Post-Hoc-Prob denotes the post-hoc deferral rule which includes the entire probability vector padded with zeros. 
    Post-Hoc-SortedProb denotes the post-hoc deferral rule which includes the sorted probability vector padded with zeros.
    }
    \vspace{-2mm}
    \label{fig:feature_ablations}
\end{figure}

%

%

%
\subsection{Output length and BLEURT correlation}

Table~\ref{tbl:wmt_bleurt_length_correlation}
shows there is a significant (negative) correlation of BLEURT scores with prediction lengths across FLAN T5 model sizes.
This verifies there is some signal in using output length as a criterion for deferral.

\begin{table}[!t]
    \small
    \centering
\begin{tabular}{@{}llllllll@{}}
\toprule
\toprule
   {Dataset} & {Small} &     {Base} & {Large} & {XL} & {XXL} \\
\midrule
 WMT DE $\to$ FR &  $-0.35$ &        $-0.28$ &        $-0.22$ &           $-0.30$ &             $-0.24$ &             \\
 WMT EN $\to$ FR &  $-0.26$ &        $-0.58$ &        $-0.26$ &           $-0.33$ &             $-0.29$ &             \\
 WMT FR $\to$ EN &  $-0.16$ &        $-0.20$ &        $-0.21$ &           $-0.24$ &             $-0.10$ &             \\
\bottomrule
\end{tabular}
    \caption{Correlation of BLEURT scores with prediction lengths across FLAN-T5 model sizes. 
    We observe that there is a non-zero correlation between output lengths and BLEURT scores. 
    Hence, there is some signal in using output length as a criterion for deferral.
    }
    \label{tbl:wmt_bleurt_length_correlation}
\end{table}

\begin{figure}[!t]
    \centering
    \includegraphics[width=0.625\linewidth]{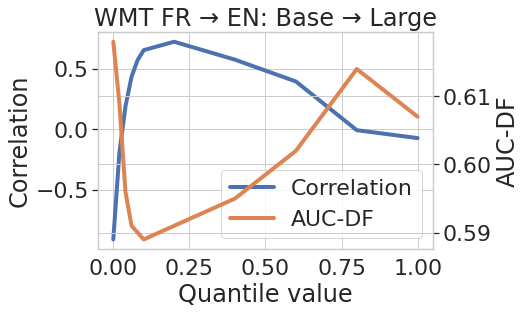}
    \caption{Different quantile scores have different correlations with prediction lengths and different AUC-DF for the deferral curves. 
     For this particular dataset, the best quantile values are 0.0 and 0.8.
     }
    \label{fig:correlation_auc}
\end{figure}


%
\subsection{Additional results for various models and datasets}

\begin{figure}[!t]
    \centering
    \includegraphics[width=1.0\linewidth]{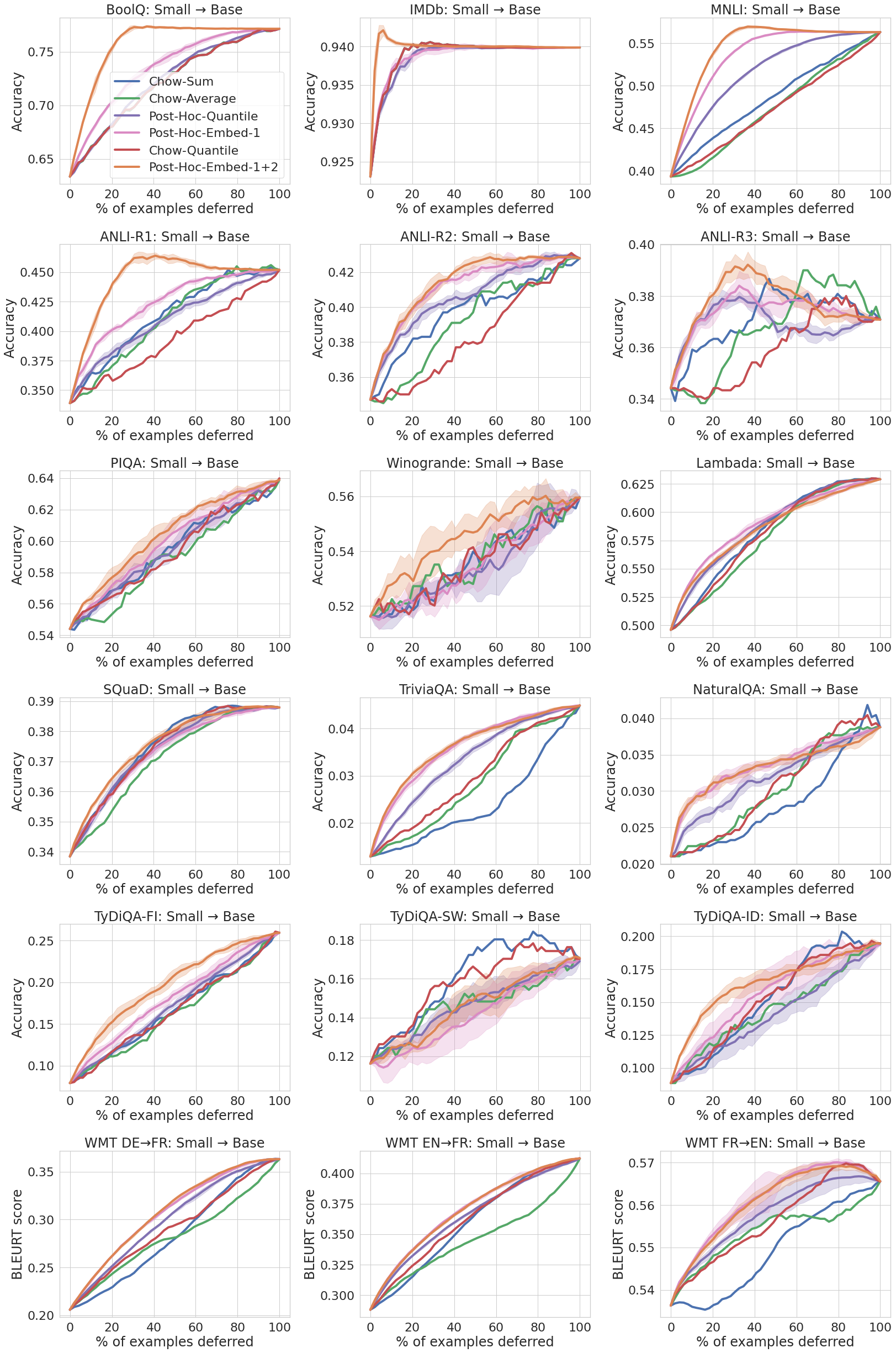}
    \caption{Cascade curves for all datasets and deferral methods for Small $\to$ Base model.
     }
\end{figure}

\begin{figure}[!t]
    \centering
    \includegraphics[width=1.0\linewidth]{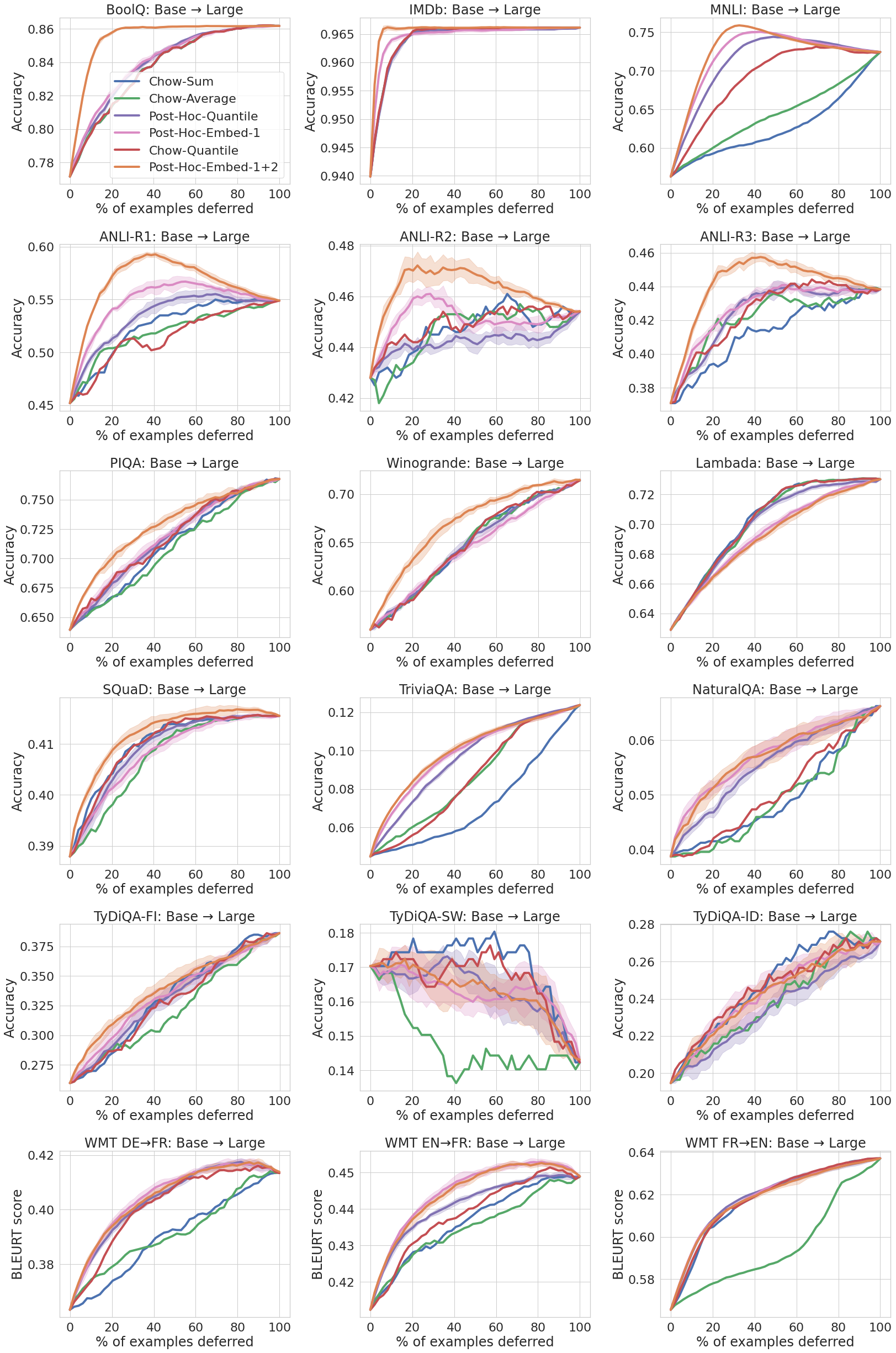}
    \caption{Cascade curves for all datasets and deferral methods for Base $\to$ Large model.
     }
\end{figure}

\begin{figure}[!t]
    \centering
    \includegraphics[width=1.0\linewidth]{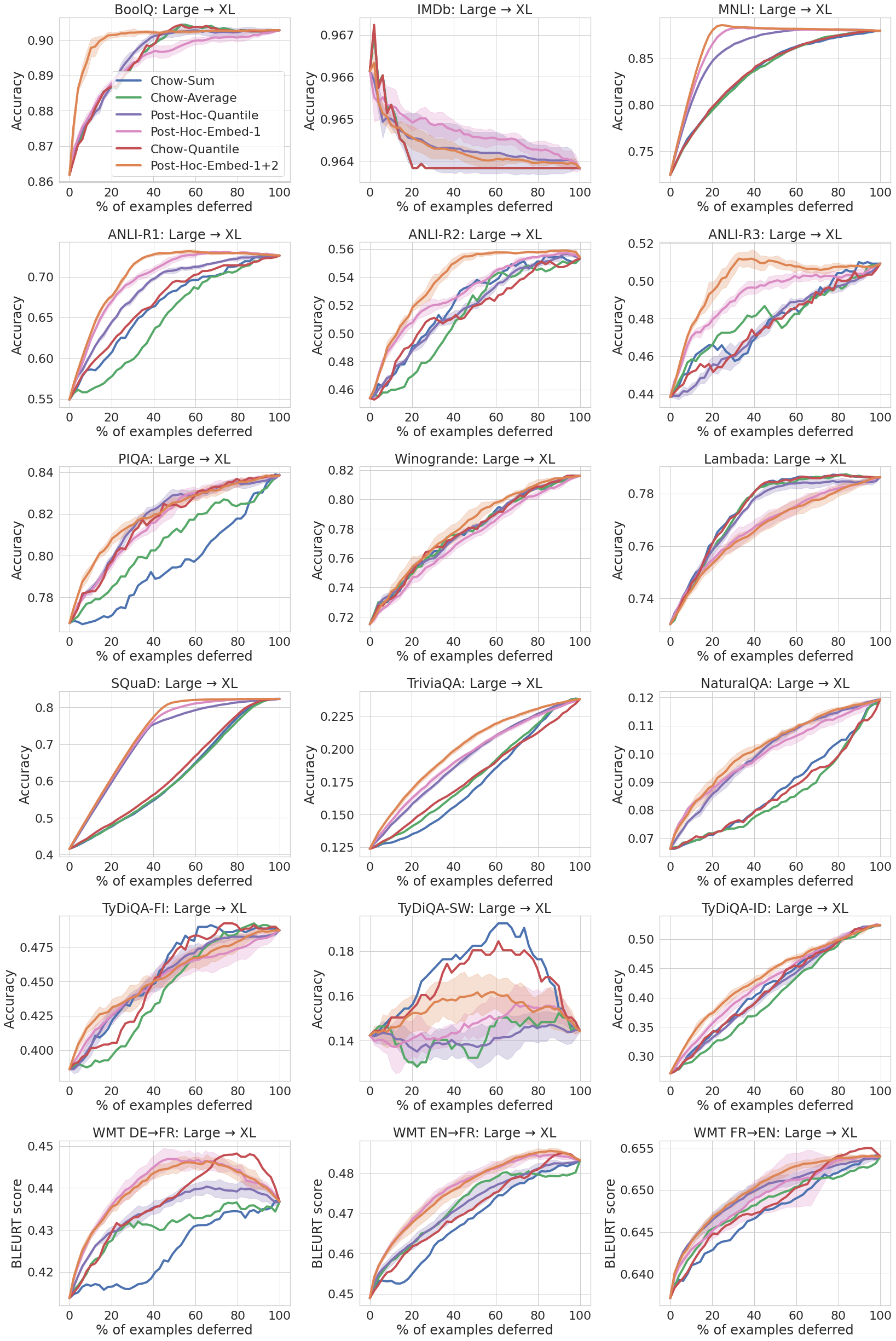}
    \caption{Cascade curves for all datasets and deferral methods for Large $\to$ XL model.
     }
\end{figure}
\clearpage

\subsection{Different version of length plot}
\begin{figure*}[h]
    \centering
    %
    %
    \begin{subfigure}{0.3\textwidth}
    \centering
     \includegraphics[width=\linewidth]{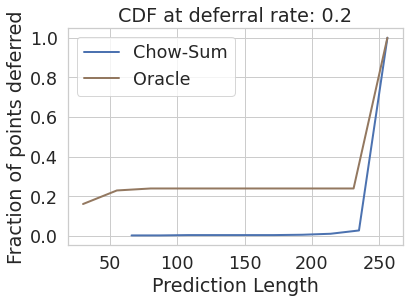}
     \caption{}
    \end{subfigure}
    \begin{subfigure}{0.3\textwidth}
    \centering
     \includegraphics[width=\linewidth]{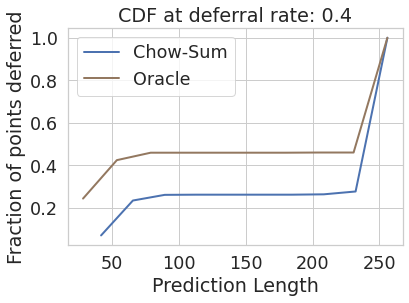}
     \caption{}
    \end{subfigure}
    \begin{subfigure}{0.3\textwidth}
    \centering
     \includegraphics[width=\linewidth]{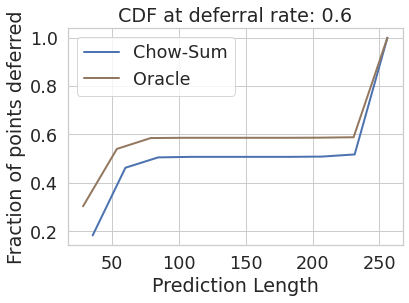}    
     \caption{}
    \end{subfigure}
     \caption{CDF plot for the  WMT FR $\to$ EN dataset for the FLAN T5 Base $\to$ Large model for different deferral rates. Each plot considers the points which are deferred by each method at the mentioned deferral rate. Each point $(x,y)$ represents that $y$ fraction of points at the desired deferral rate have prediction lengths less than $x$. We can see that Oracle has larger fractions of points deferred with smaller prediction length. Hence, Chow-Sum prefers to defer longer predictions as compared to the Oracle method.}
\end{figure*}

\begin{figure*}[h]
    \centering
    %
    %
    \begin{subfigure}{0.3\textwidth}
    \centering
     \includegraphics[width=\linewidth]{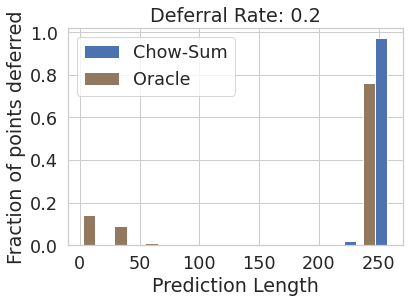}
     \caption{}
    \end{subfigure}
    \begin{subfigure}{0.3\textwidth}
    \centering
     \includegraphics[width=\linewidth]{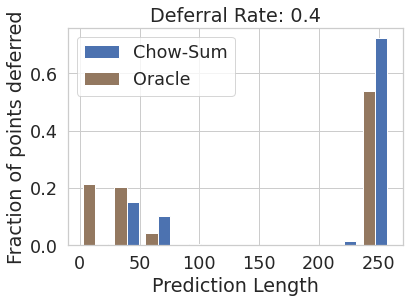}
     \caption{}
    \end{subfigure}
    \begin{subfigure}{0.3\textwidth}
    \centering
     \includegraphics[width=\linewidth]{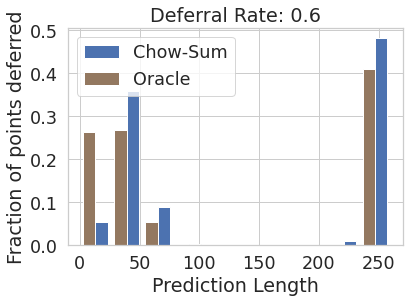}    
     \caption{}
    \end{subfigure}
     \caption{CDF plot for the  WMT FR $\to$ EN dataset for the FLAN T5 Base $\to$ Large model. Each plot considers the points which are deferred by each method at the mentioned deferral rate. We can see that Oracle has larger fractions of points deferred with smaller prediction lengths. Hence, Chow-Sum prefers to defer longer predictions as compared to the Oracle method.}
\end{figure*}
\clearpage

\subsection{Cascade curves with relative latency}

\begin{figure*}[h]

    \resizebox{0.99\linewidth}{!}{%
        \includegraphics[scale=0.3]{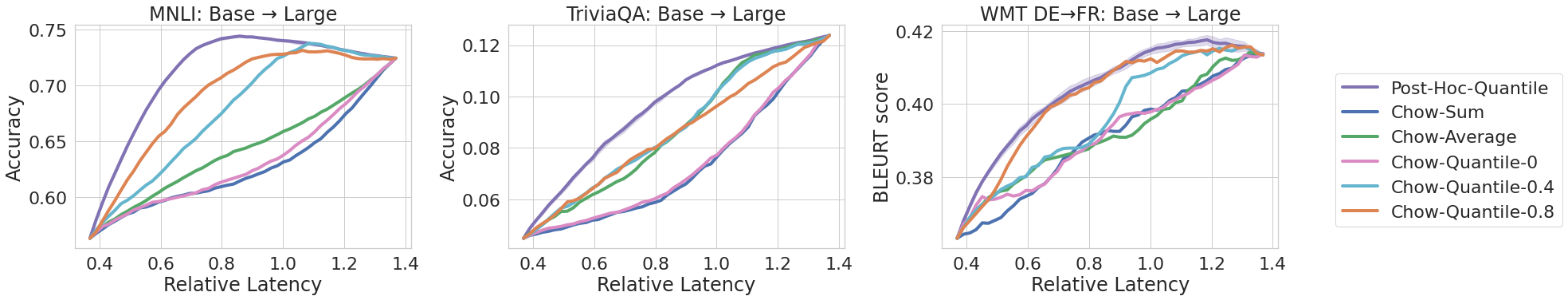}
    }
    
    \caption{Deferral curves on MNLI, TriviaQA, and WMT DE $\to$ FR for a FLAN-T5 Base $\to$ Large cascade.}

\end{figure*}

\clearpage
\subsection{Results with encoder and decoder embeddings}

Post-Hoc-Embed-Dec-1 is the method Post-Hoc-Embed-1 method from main draft which includes decoder embeddings from model 1. Post-Hoc-Embed-Enc+Dec-1 additionally includes encoder embeddings from model 1. 

Similarly, Post-Hoc-Embed-Dec-1+2 is the method Post-Hoc-Embed-1+2 method from main draft which includes decoder embeddings from model 1 and intermediate decoder embeddings from model 1. Post-Hoc-Embed-Enc+Dec-1+2 additionally includes encoder embeddings from model 2.

We do not observe any additional gains from including encoder embeddings in these methods.

\begin{figure*}[h]

    \resizebox{0.99\linewidth}{!}{%
        \includegraphics[scale=0.3]{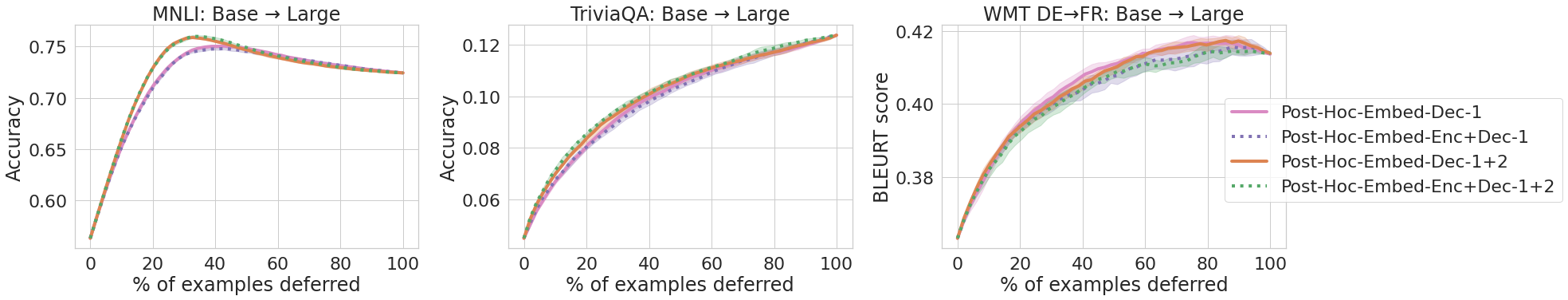}
    }
    
    \caption{Deferral curves on MNLI, TriviaQA, and WMT DE $\to$ FR for a FLAN-T5 Base $\to$ Large cascade with additional encoder embeddings.}

\end{figure*}

\clearpage
\subsection{Results with beam search decoding}

\begin{figure*}[h]

    \resizebox{0.99\linewidth}{!}{%
        \includegraphics[scale=0.3]{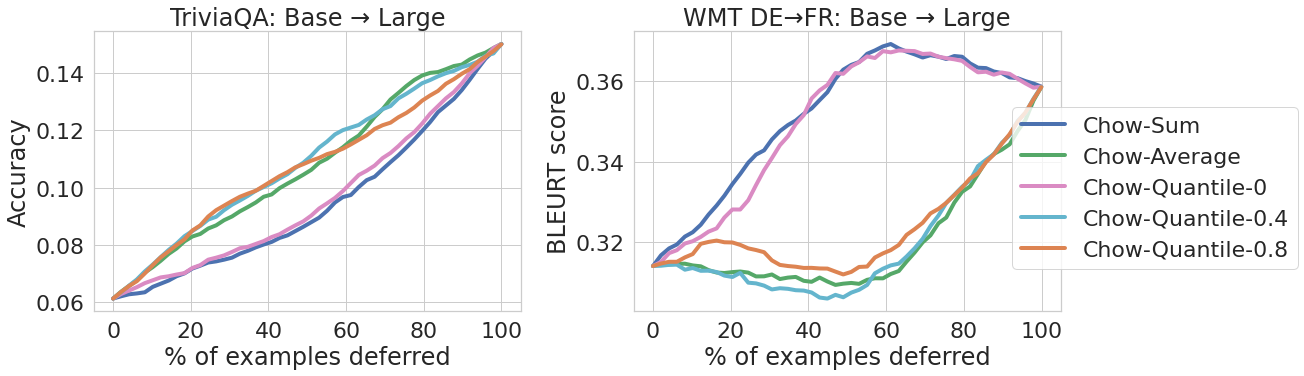}
    }
    \caption{Deferral curves on TriviaQA, and WMT DE $\to$ FR for a FLAN-T5 Base $\to$ Large cascade with beam search decoding.}
\end{figure*}

\begin{figure*}[h]

    \resizebox{0.99\linewidth}{!}{%
        \includegraphics[scale=0.3]{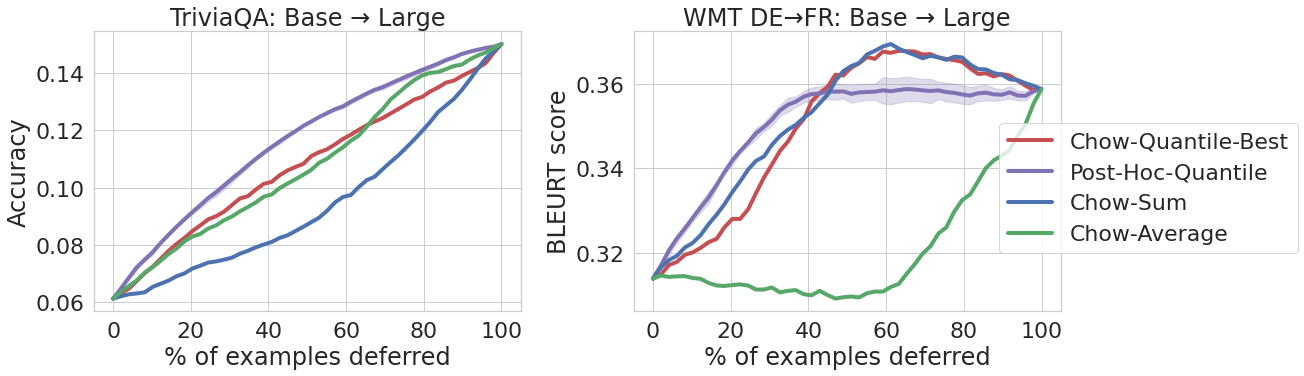}
    }
    \caption{Deferral curves on TriviaQA, and WMT DE $\to$ FR for a FLAN-T5 Base $\to$ Large cascade with beam search decoding.}

\end{figure*}

\begin{figure*}[h]
    \centering
    %
    %
    \begin{subfigure}{0.48\textwidth}
    \centering
     \includegraphics[width=\linewidth]{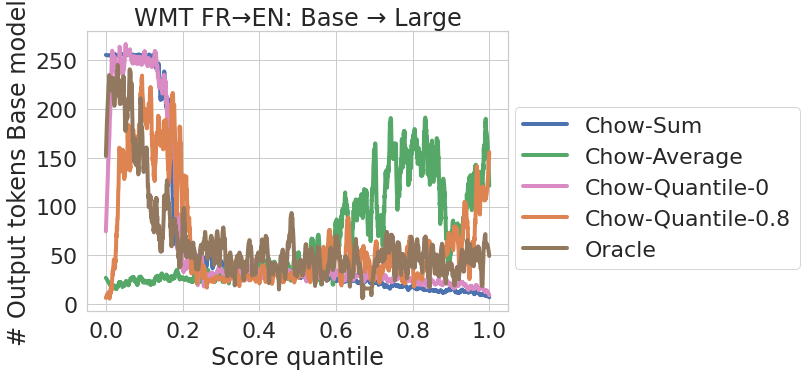}
     \caption{}
    \end{subfigure}
    \begin{subfigure}{0.48\textwidth}
    \centering
     \includegraphics[width=\linewidth]{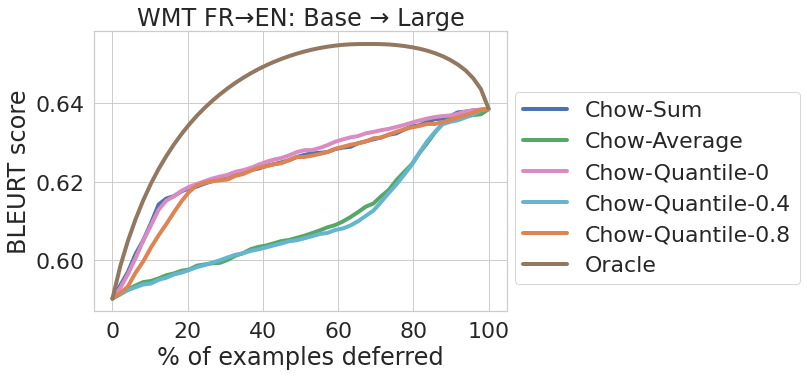}
     \caption{}
    \end{subfigure}

     \caption{{\bf Left}:
Relation between deferral rules and output length (number of tokens) for WMT FR $\rightarrow$ EN dataset and FLAN-T5 Base Model with decoding using beam search. 
{\tt Chow-Sum} tends to defer longer prompts:
the prompts with lowest scores have notably higher length than those with higher scores.
Interestingly, {\tt Chow-Average} \emph{over-}corrects this bias:
it tends to overly defer prompts with \emph{lower} length. Oracle refers to deferring using the difference of BLEURT scores of the two models. 
Oracle also tends to defer longer outputs, but the preference is moderate as compared to {\tt Chow-Sum}. 
{\bf Right}: Corresponding deferral curves. }
\end{figure*}

\begin{figure*}[h]

    \resizebox{0.99\linewidth}{!}{%
        \includegraphics[scale=0.3]{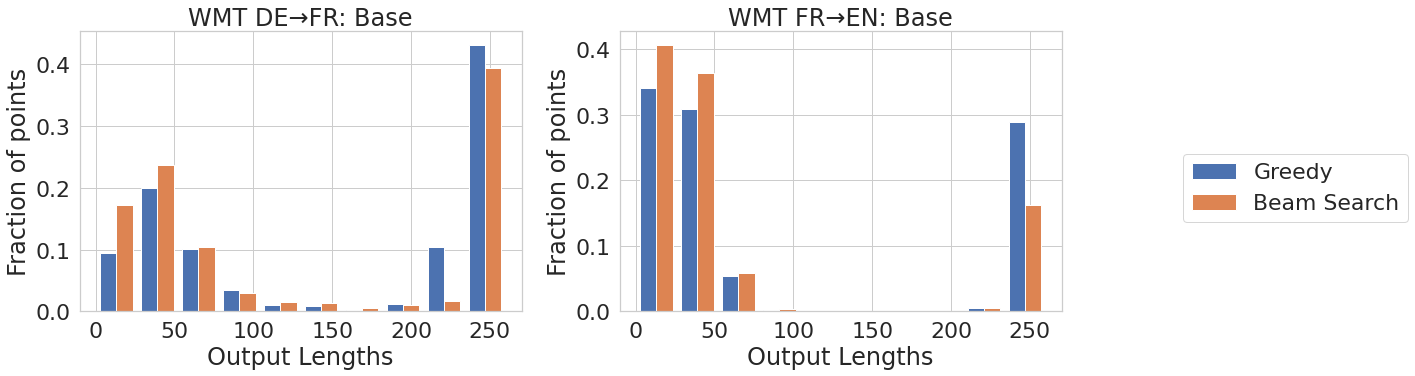}
    }
    \caption{Comparison of Output lengths on  WMT DE $\to$ FR and WMT FR $\to$ EN for a FLAN-T5 Base model with beam search decoding vs. greedy decoding. We observe that greedy decoding does seem to have overall shorter output lengths as compared to greedy decoding but still puts a considerable weight on longer outputs.}

\end{figure*}

\subsection{Performance of methods in predicting golden deferral label}

\begin{table}[h]
    \centering
    
    \resizebox{\linewidth}{!}{
\begin{tabular}{@{}lccccccc@{}}
\toprule
\toprule
   {\bf Dataset} &    {\bf\tt Chow-Sum} & {\tt Chow-Average} & {\bf\tt Chow-Quantile-Best} & {\bf\tt Post-Hoc-Quantile}&     {\bf\tt Post-Hoc-Embed-1}&     {\bf\tt Post-Hoc-Embed-1+2} \\
\midrule
      MNLI  &       0.46 &        0.53 &           0.73 &             0.86 &  0.92 & 0.95  \\

  TriviaQA  &        0.31 &       0.53 &           0.52 &             0.68 &             0.70 & 0.71  \\

 WMT DE $\rightarrow$ FR &       0.51 & 0.56 &  0.62 &             0.62 &             0.63 & 0.62  \\
\bottomrule
\end{tabular}
}
    \caption{The table shows the AUC-ROC score for predicting the golden binary deferring labels, which is 1 if model 2 (FLAN-T5 Large) is better than model 1 (FLAN-T5 Base), and 0 otherwise. We see that {\tt Chow-Sum} and {\tt Chow-Average} tend to be poorly predictive of this golden label. The recommended {\tt Post-Hoc} deferral approaches yield significant gains in the AUC-ROC.
    }
    \label{tbl:area_under_curve}
\end{table}

\clearpage

\section{How is $\Phi( \cdot )$ computed?}
\label{phi-setup}

Here, we give an example of how $\Phi(\boldsymbol{x})$ is computed for each of the Post-Hoc methods. 

For concreteness, suppose that the example $\boldsymbol{x}$ outputs a prediction which has 5 output tokens. Let the output log probability associated with each output token be $p_i(\boldsymbol{x})$ where $i \in \{ 0, 1, 2, 3, 4 \}$. Then, the Chow-Sum score is $s_{\rm sum}(\boldsymbol{x}) = \sum_i p_i(\boldsymbol{x})$ and the Chow-Average score is $s_{\rm avg}(\boldsymbol{x}) = \frac{1}{5}\cdot\sum_i p_i(\boldsymbol{x})$. Let the various quantiles be $s_{\rm quant}( \boldsymbol{x}, \alpha)$ where $\alpha \in [0, 0.01, 0.02,... 0.1, 0.2, 0.3, …, 1.0]$. For concreteness, we take 20 quantiles. 

Let the decoder embedding from model 1 averaged over all output tokens be denoted as ${\rm Emb}_{\rm dec, \rm final, \rm avg,\rm m1}(\boldsymbol{x}) \in \Real^{n_1}$. 
Let the decoder embedding from model 2’s first layer for the first output token be denoted as ${\rm Emb}_{\rm dec,\rm int,\rm first, \rm m2}(\boldsymbol{x}) \in \Real^{n_2}$. 

\textbf{Post-Hoc-Quantile}
For the method Post-Hoc-Quantile, 
$$\Phi(\boldsymbol{x}) = [ {s_{\rm sum}(\boldsymbol{x}), s_{\rm avg}(\boldsymbol{x}), s_{\rm quant}( \boldsymbol{x}, \alpha_i)} ]$$
where we concatenate the quantities, Chow-Sum and Chow-Average scores which makes this feature 22 dimensional. 

\textbf{Post-Hoc-Embed-1}
For the method Post-Hoc-Embed-1, 
$$\Phi(\boldsymbol{x}) = [ {s_{\rm sum}(\boldsymbol{x}), s_{\rm avg}(\boldsymbol{x}), s_{\rm quant}(x, \alpha_i), {\rm Emb}_{\rm dec, \rm final, \rm avg,\rm m1}(\boldsymbol{x})} ]$$ 
where we concatenate the quantities, Chow-Sum, Chow-Average scores and model 1’s decoder embeddings which makes this feature $22+n_1$ dimensional. 

\textbf{Post-Hoc-Embed-1+2}
For the method Post-Hoc-Embed-1+2, 
$$\Phi(\boldsymbol{x}) = [ {s_{\rm sum}(\boldsymbol{x}), s_{\rm avg}(\boldsymbol{x}), s_{\rm quant}( \boldsymbol{x}, \alpha_i), {\rm Emb}_{\rm dec, \rm final, \rm avg,\rm m1}(\boldsymbol{x}), {\rm Emb}_{\rm dec,\rm int,\rm first, \rm m2}(\boldsymbol{x})} ]$$ 
where we concatenate the quantities, Chow-Sum, Chow-Average scores, model 1’s decoder embeddings and model 2’s intermediate decoder embeddings  which makes this feature $22+n_1+n_2$ dimensional. 


\end{document}